\documentclass[letterpaper, 10 pt, conference]{ieeeconf} 
\usepackage{amsmath,amssymb}
\usepackage{afterpage}
\usepackage{graphicx}
\usepackage{float}
\IEEEoverridecommandlockouts
\usepackage{cite}
\usepackage{pstricks}
\usepackage{comment}
\usepackage{algorithm}
\usepackage[noend]{algpseudocode}
\renewcommand{\baselinestretch}{0.99}
\usepackage{graphicx}

\title{\LARGE \bf Adversarial Learning-Based On-Line Anomaly Monitoring\\ for Assured Autonomy}

\author{Naman Patel$^{*, 1}$, Apoorva Nandini Saridena$^{*,2}$, Anna Choromanska$^{*,2}$, Prashanth Krishnamurthy$^{*,1}$, Farshad Khorrami$^{*,1}$
\thanks{$^{*}$All authors contributed equally and they belong to the $^{1}$Control/Robotics Research Laboratory (CRRL) and the $^{2}$Machine Learning Laboratory in the Department of Electrical and Computer Engineering, NYU Tandon School of Engineering.
{\tt\small \{naman.patel, ans609, ac5455, prashanth.krishnamurthy, khorrami\}@nyu.edu.} This work was supported in part by the U.S. Office of Naval Research under Award N00014-15-1-2182.}%
}

\begin{document}

\IEEEoverridecommandlockouts
\pubid{\makebox[\columnwidth]{978-1-5386-5541-2/18/\$31.00~\copyright2018 IEEE \hfill} \hspace{\columnsep}\makebox[\columnwidth]{ }}
\maketitle
\pubidadjcol

\begin{abstract}

The paper proposes an on-line monitoring framework for continuous real-time safety/security in learning-based control systems (specifically application to a unmanned ground vehicle). We monitor validity of mappings from sensor inputs to actuator commands, controller-focused anomaly detection (CFAM), and from actuator commands to sensor inputs, system-focused anomaly detection (SFAM). CFAM is an image conditioned energy based generative adversarial network (EBGAN) in which the energy based discriminator distinguishes between proper and anomalous actuator commands.  SFAM is based on an action condition video prediction framework to detect anomalies between predicted and observed temporal evolution of sensor data. We demonstrate the effectiveness of the approach on our autonomous ground vehicle for indoor environments and on Udacity dataset for outdoor environments.

\end{abstract}

\section{INTRODUCTION}
\label{sec:intro}

The development of algorithms for autonomous navigation of robotic systems in complex unstructured environments is attracting significant research interest due to emergence of more accurate and cheaper sensors, faster embedded computing devices (e.g., GPUs), and vast amounts of data. Particularly, learning-based algorithms are widely used due to the large sets of available data to train either the perception subsystems or the end-to-end control policy for robust autonomous operation of the robotic system \cite{bojarskiTDFFGJM16,patel2017b}. 

These learning-based systems cannot be trained in all possible environments and environmental conditions. Such systems are also vulnerable to adversarial attacks \cite{Eykholt_2018_CVPR,patel2018b}. Recently, approaches have been proposed to address this challenge based on generative modeling \cite{LiangLS18,LeeLLS18} and gradient based search \cite{tian2018deeptest}. Formal verification \cite{KatzBDJK17} and Bayesian optimization \cite{Ghosh2018Verifying} are promising but are computationally complex.

\begin{figure}[h!tbp]
    \centerline{\includegraphics[width = 0.8\columnwidth]{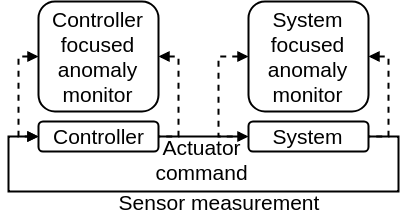}}
    \caption{The proposed learning-based anomaly detection framework for continuous real-time monitoring of an autonomous closed-loop system \cite{khorramiKK16}.}
    \label{fig:framework}
\end{figure}

For real-time detection of anomalies/malfunctions in a cyber-physical system \cite{KelirisSCKMK16,AmrouchKPHKK17}, we propose the framework in Figure~\ref{fig:framework} which combines complementary anomaly monitoring methods \cite{khorramiKK16,KrishnamurthyKKS18}, namely controller-focused anomaly monitor (CFAM) and system-focused anomaly monitor (SFAM). CFAM uses an image-conditioned EBGAN to validate the mapping from the sensor data to the actuator command. SFAM uses an action conditioned video prediction system to validate the mapping from the system action to the sensor data. The proposed learning-based anomaly detection framework continuously validates and verifies autonomous operation during run-time. Novel aspects of this study are:
\begin{itemize}
\item A learning-based on-line framework to monitor the controller and the system behavior.
\item A conditional EBGAN architecture to detect anomalies in controller outputs.
\item An action conditioned video prediction framework to detect anomalies in system behavior.
\item A methodology to train a robust video prediction architecture to detect anomalies. 
\end{itemize}

This paper is organized as follows. The background and the problem are discussed in Section~\ref{sec:problem_formulation}. The architecture, training methodology, and anomaly detection for CFAM and SFAM are described in Section~\ref{sec:proposed_framework}. Section~\ref{sec:exp_studies} reports results on indoor (collected from our experimental unmanned ground vehicle) and outdoor (Udacity \cite{udacitydata}) datasets. Section~\ref{sec:conclusion} concludes the paper.

\section{Problem Formulation}
\label{sec:problem_formulation}

\begin{figure*}[!htbp]
    \centerline{\includegraphics[width = 0.95\linewidth]{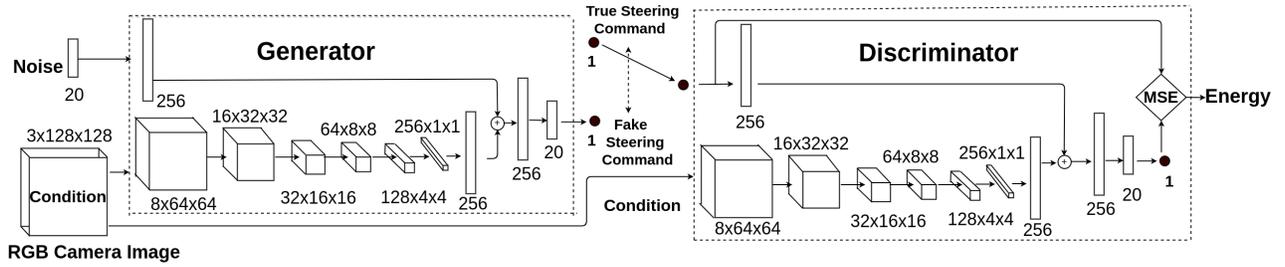}}
    \caption{Conditional energy based generative adversarial network (CEBGAN) framework for CFAM.}
    
    \label{fig:cEBGAN}
\end{figure*}

The presented anomaly detection applies to any closed-loop system such as an autonomous vehicle. In this paper, we consider an unmanned ground vehicle (UGV) instrumented with camera(s) and LIDAR and controlled by an end-to-end learning system. The input command to the vehicle is $u_{t}$ at time step $t$ with the corresponding sensor output, $x_{t}$.

The CFAM operates on the stream of sensor data and actuator commands to monitor validity of the mapping $x\rightarrow u$ and detects anomalies related to the controller. These anomalies can be due to distributional shift (e.g., when the testing and training environments are different) or due to sensor malfunctions/failures/attacks. The approach for CFAM is based on  learning sensor data (image) conditioned energy based generative adversarial network (CEBGAN) in Figure~\ref{fig:cEBGAN} which outputs low energy for valid actuator (steering) commands and high energy for anomalous commands. 

The SFAM monitors the validity of the (dynamic) mapping $u\rightarrow x$ from system action to sensor data (over a time horizon into future) and detects anomalies generated due to malfunctions in the system. Malfunctions/anomalies can occur due to environmental perturbation (e.g., slippery road) or partial/full subsystem failure (e.g., electrical, mechanical) which causes the system to not act according to the given actuator command. The framework addresses this problem by learning the dynamics of the overall system based on action conditioned video prediction as shown in Figure~\ref{fig:videopredgen}. 
In the implementation, the framework takes as input $x_{t-3:t}$ and actuator commands $u_{t-2:t+1}$ to predict $x_{t+1}$, which is used to validate the system action to system data mapping. The video prediction and anomaly detection architectures for the proposed framework are shown in Figures~\ref{fig:videopredanom_arch} - \ref{fig:videopredgan}.

\section{Proposed Framework}
\label{sec:proposed_framework}

\subsection{Image conditioned EBGAN}
\subsubsection{Model architecture}
\label{sec:cfam_arch}
Motivated by the success of GANs\cite{Goodfellow2014} in a variety of applications, we propose to use discriminator in GAN for CFAM. We combine conditional-GAN\cite{MirzaO14} with EBGAN\cite{Zhao2017}, to generate CEBGAN and explore its potential to detect controller generated anomalies.

Inputs to the generator are noise vector $z$ sampled uniformly from $(0,1)$ and condition image $x$. Discriminator inputs are actuator command $u$ and condition image $x$. The discriminator learns to map inputs $u$ and $x$ to a scalar value (Energy), while the generator learns to predict steering command from the condition image.The discriminator output is computed as mean square error between input steering command $u$ and the condition $x$ driven steering command prediction.  For CFAM, the discriminator is the key component and the generator is merely trained to produce contrastive samples. We train them both simultaneously and use only the discriminator in CFAM.

In CEBGAN, the generator and the discriminator are conditioned on camera input images. The conditional image $x$ feature vector is computed by passing the image through a series of convolutional layers. In our experiments with indoor (UGV) and outdoor (Udacity) datasets, the images are all first rescaled to size 3x128x128 and the best performer discriminator used 6 convolution layers for image feature extraction in both generator and discriminator. The number of kernels in the first, second, third, fourth, fifth and sixth convolution layers are 8, 16, 32, 64, 128, 256 respectively and kernels size is fixed to 4x4  and applied with stride 2. Before each convolution layer, spatial batch normalization is performed and  LeakyReLU activations with 0.2 negative slope are used for all layers except for the output, which uses Tanh.The obtained feature map is then reshaped as a vector.

In generator, the noise input $z$ is mapped to a fully connected hidden layer of size same as that of the feature vector computed on the condition image to allow their summation. Then, in both, the summation of these vectors is mapped to another fully connected hidden layer before it is reduced to a single output neuron as shown in Figure~\ref{fig:cEBGAN}.


\subsubsection{Training methodology}
\label{sec:CFAM_training}

In CEBGAN, the discriminator is trained with an objective function in order to shape the energy function to attribute low energies to correct steering commands and high energies to the generated (or anomalous) commands. We use different loss functions to train the discriminator and the generator.
Given a positive margin $m$, a condition image $x$,  true steering command  $u$, and generated steering command $G(z)$, the discriminator loss $L_{D}$ and generator loss $L_{G}$ are defined as:
\begin{eqnarray}
L_{D}(u, z, x) &=& D(u, x) + [m - D(G(z), x)]^{+} \\
L_{G}(z, x) &=& D(G(z), x)
\end{eqnarray}
where $D{(u, x)}$ is discriminator output and $[.]^{+} = max(0, .)$.
The generator and discriminator parameters are optimized using Adam optimizer.

\subsubsection{Anomaly detection framework}
\label{sec:CFAM_anomalydetection}
The real-time CFAM module consists of a discriminator trained as described in Section~\ref{sec:CFAM_training}. Given a condition image $x_t$, energies (discriminator outputs) are computed for $\textbf{N}$ linearly spaced equi-distant points in the steering commands values range. The validity of the incoming sensor data is determined by computing the steering command deviation as the magnitude of the difference of the controller output (steering command) and the corresponding minimum energy steering command value. Anomalies are associated with data points with deviation greater than the threshold.


\subsection{Adversarially learned action conditioned video prediction}

\begin{figure}[!htbp]
\includegraphics[width = \linewidth]{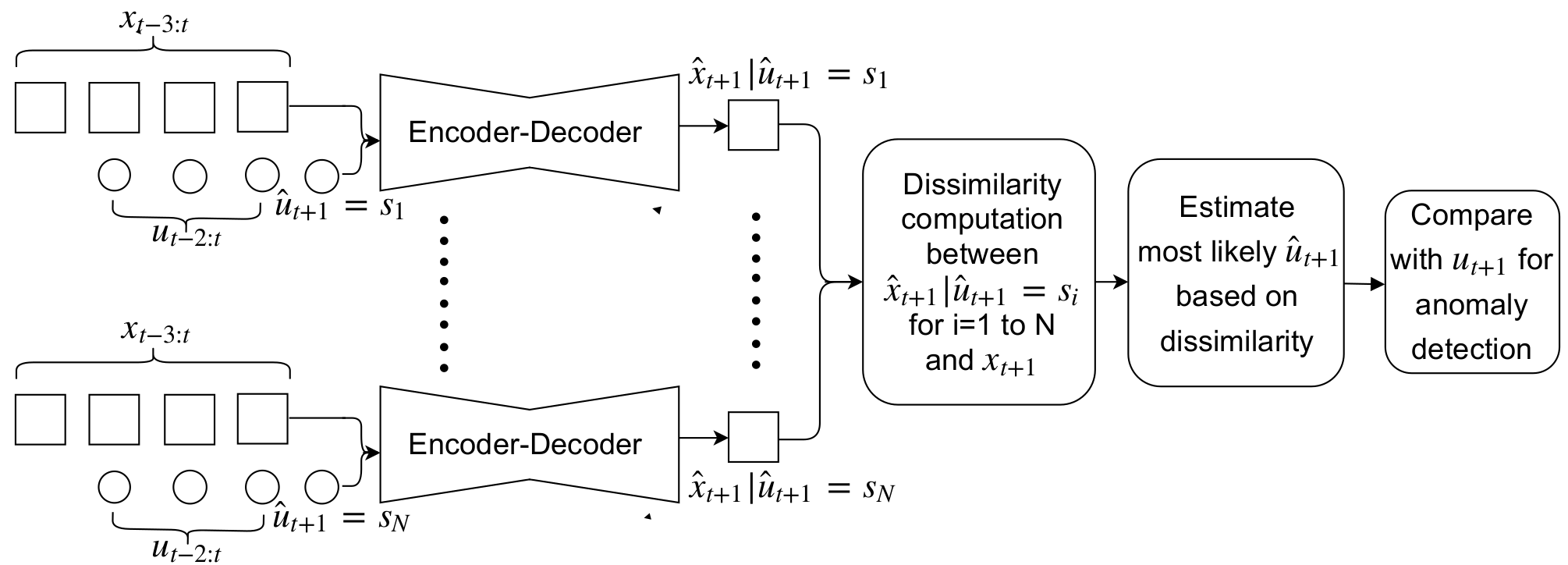}
    \caption{Video prediction based SFAM: It consists of a video prediction generator architecture, dissimilarity computation, and validation modules.}
    \label{fig:videopredanom_arch}
\end{figure}

\begin{figure*}[!tpbh]
    \includegraphics[width = \linewidth]{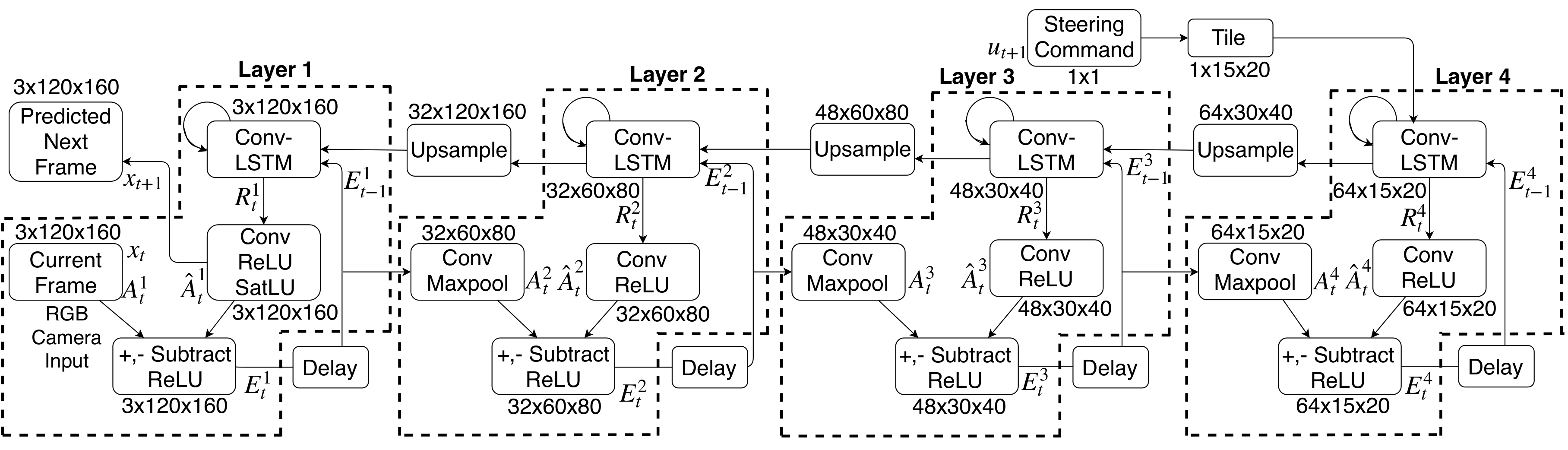}
    \caption{Deep predictive coding based action conditioned video prediction architecture.}
    \label{fig:videopredgen}
\end{figure*}
\subsubsection{Model architecture}
\label{sec:sfam_arch}
\begin{figure}[!htbp]
    \centerline{\includegraphics[width = \linewidth]{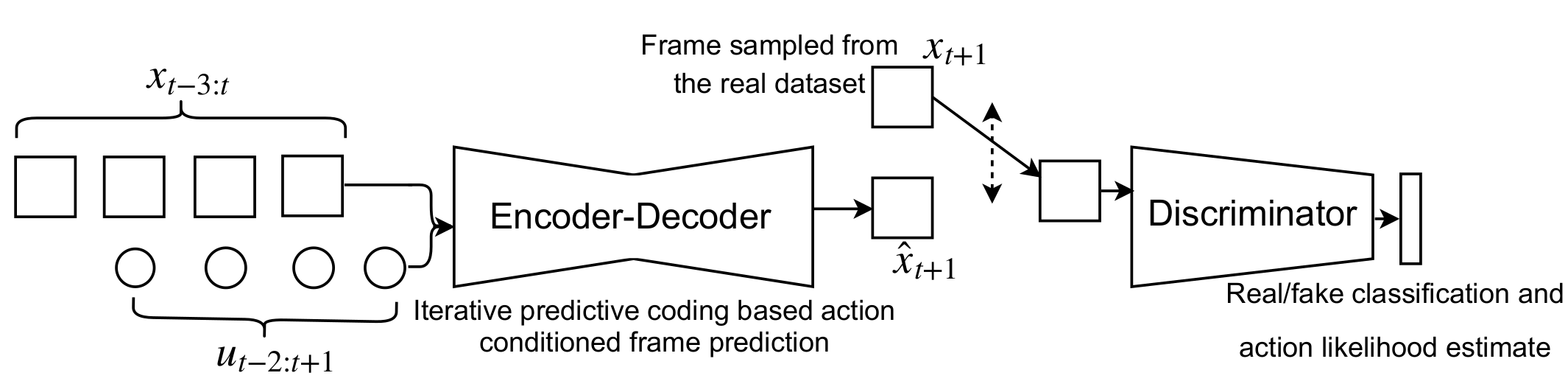}}
    \caption{
Adversarial training architecture for video prediction based on GAN for SFAM.}
    \label{fig:videopredgan}
\end{figure}

Learning-based video prediction has been used in various control applications \cite{SrivastavaMS15,oh2015action,FinnL17,BabaeizadehFECL18}. The SFAM in Figure~\ref{fig:videopredanom_arch} consists of an action conditioned video prediction architecture shown in Figure~\ref{fig:videopredgen}, dissimilarity computation module, and validation module. The video prediction architecture is based on the four-layer Prednet \cite{LotterGC2017}. Each layer has four sub-modules: (1) convolution, (2) prediction representation, (3) recurrent representation, and (4) error representation.

All convolutions have 3x3 kernels and max-pooling of stride 2x2 with kernel size 2x2. The number of output convolution channels per layer for prediction and target representation sub-modules are 3, 32, 48, and 64, respectively. 

The inputs to the network are a sequence of images $x_{t}$ and the steering commands $u_{t+1}$ at the next time step as shown in Figure~\ref{fig:videopredgan}. Each layer $l$ at time step $t$ has prediction $\hat{A}^{l}_{t}$ and target representations $A^{l}_{t}$, generated by the prediction and convolution sub-modules. At the first layer, the target representation $A^{l}_{t}$ is the actual frame. In layers 2, 3, 4 the target representation is generated by the convolution sub-module with the error representation from the previous layer $E^{l-1}_{t}$. The convolution sub-module has a convolution layer with ReLU nonlinearity followed by max-pooling layer. 

The prediction representation $\hat{A}^{l}_{t}$ is generated from the recurrent representation of the current layer $R^{l}_{t}$ as input to the prediction sub-module. The prediction module consists of a convolution layer with ReLU non-linearity except for at the first layer where it is followed by a SatLU non-linearity to saturate the values to the maximum pixel intensity values. 

The recurrent representation layer that generates $R^{l}_{t}$ is a convolutional LSTM whose inputs are the error representation from the previous time step of the same layer $E^{l}_{t-1}$ combined with the upsampled output of the recurrent representation of the higher layer $R^{l+1}_{t}$. The hidden state is the recurrent representation from prior time step $R^{l}_{t-1}$. 

The error representation $E^{l}_{t}$ is generated by combining the feature maps of the difference between the prediction $\hat{A}^{l}_{t}$ and the target $A^{l}_{t}$ and the target $A^{l}_{t}$ and the prediction $\hat{A}^{l}_{t}$ followed by a ReLU nonlinearity. In the first time step, all error representations are reset to zero. The steering command is introduced at the next time step $u_{t+1}$ as the input to generate action conditioned frame prediction. It is concatenated with the error representation of the fourth layer by tiling the steering command to the same dimension and used as input to the fourth recurrent representation layer.

\subsubsection{Training methodology}
\label{sec:sfam_train_methodology}
The video prediction architecture takes as input four RGB images $x_{t-3:t}$ (scaled between 0 and 1) from a monocular camera of resolution 120x160 and steering commands $u_{t-2:t+1}$ to predict the frame $\hat{x}_{t+1}$. The training has two stages: error representation minimization and adversarial optimization. The weights are updated using an Adam optimizer with a base learning rate of 0.001. 

The loss function for the error representation minimization consists of the average of all error representation $E^{l}_{t}$ at each time step $t$ and at each layer $l$, the negative of structural similarity (SSIM) (kernel size=5) between the predicted frame $\hat{x}_{t+1}$ and the actual frame $x_{t+1}$ and the SSIM between the predicted frame $\hat{x}_{t+1}$ and the previous frame $x_{t}$. The weight for the average of all error representation is 0.1. The weight for structural similarity between actual and predicted frame is 1. The weight for structural similarity between previous and predicted frame is 0.5.

Adversarial optimization is introduced to curtail blurry prediction images \cite{mathieucy16}. In this stage, the video prediction architecture is the generator and is combined with a spectrally normalized discriminator \cite{miyatoTMY18} with N+1 labels where one label is the fake/real label and the other labels are N steering commands as in Figure~\ref{fig:videopredgan}.  We use 15 equally spaced labels for steering command actions for both indoor and Udacity dataset with range from $-0.24$ to 0.28  and from $-0.52$ to 0.56 respectively. The discriminator consists of nine spectrally normalized convolution layers with LeakyReLU non-linearity with negative slope of 0.1 in between all of them and a kernel size alternating between 3x3 and 4x4 with stride of 1x1 and 2x2, respectively and a spectrally normalized fully connected layer, which outputs a vector of size 16. The GAN is trained based on the technique proposed in \cite{salimansGZCRCC16} with an additional regularization term as described in \cite{LeeLLS18} to make it robust to out-of-distribution samples. 

The regularization term added while updating the generator parameters is $KL(U(s)||P_{\theta}(s|G(z_{i}))$ where $U(s)$ is the uniform distribution of the steering command labels $s$, $G$ is the generator, $z_{i}$ is the video prediction network input, and $\theta$ are the parameters of the generator. This term forces the generator to output out-of-distribution samples which are in the low data density region. The generator is updated by minimizing the weighted sum of cross entropy loss (fake/real, action likelihood estimate) and the $KL$ divergence term introduced above with weights randomly generated between 0-1. 
\subsubsection{Anomaly detection framework}
\label{sec:anomaly_framework}
The video prediction framework in Figure~\ref{fig:videopredgen} generates video predictions $\hat{x}_{t+1} | u_{t+1}=s_{i}$ where $i \in [1,N]$. These video predictions are input to the dissimilarity computation module as shown in Figure~\ref{fig:videopredanom_arch} which calculates the dissimilarity ((1-SSIM)/2) of the prediction with the actual future frame. These dissimilarities are input to the validation module which selects the action corresponding to the least dissimilar predicted frame and compares it with the actual future action to validate the mapping from the actuator command (system action) to sensor data. A windowing strategy using multiple prediction frames validates system actions $\rightarrow$ sensor data mapping. 

\section{Experimental Studies}
\label{sec:exp_studies}
\subsection{Datasets}
We empirically validated CFAM and SFAM on indoor and outdoor datasets. The indoor dataset is collected by a human controller driving a UGV in a corridor with varying lighting conditions and changing obstacle placements \cite{patel2017b,patel2018a}. The Udacity dataset \cite{udacitydata} has images recorded while driving on highways and residential roads (with and without lane markings) in clear weather during daytime and includes driver's activities such as staying in and switching lanes. 

\subsection{Simulated Scenarios}
\label{sec:simulated_runs}
Datasets of RGB camera images and the corresponding steering commands for various anomalous and non-anomalous scenarios were created.  To simulate anomalous scenarios due to malfunctions/anomalies in the controller or the external system, the controller output was overridden by a human driver to create anomalous driving conditions. To test CFAM, the human-created anomalous inputs were provided to the CFAM as controller-generated commands.
To test SFAM, the actual controller-generated, correct steering commands and the sensor data that was recorded when the robotic vehicle was actually given the overridden steering commands were provided to the SFAM.

Two complementary anomalous datasets were simulated by the human driver for the UGV in the indoor environment resulting in late-right and early-left crash into a wall. In the late-right turn scenario, the driver provided steering commands to the vehicle to go straight when the vehicle has to take a right turn. In the early-left crash case, the driver provided steering commands to move towards left when the vehicle has to go straight.

\begin{figure}[!tpbh]
\centerline{
   \includegraphics[width = 0.5\linewidth]{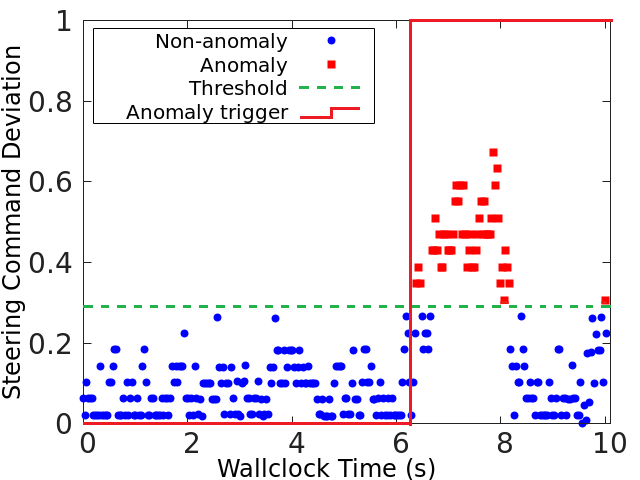}
   \includegraphics[width = 0.5\linewidth]{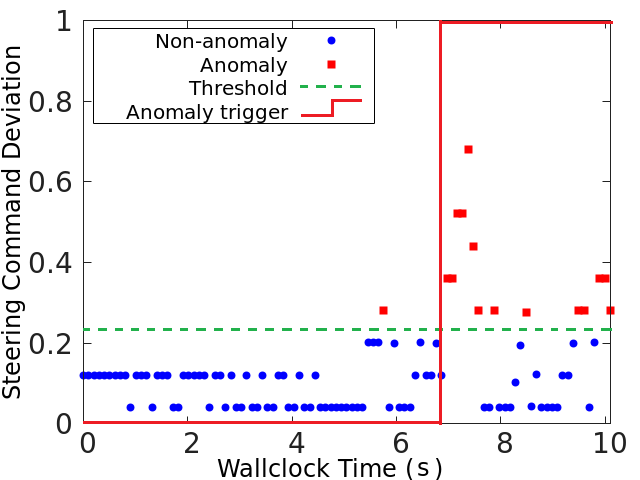} }
   \caption{Real time anomaly detection when UGV takes a late right turn. Left: controller anomaly detected by CFAM; Right: system anomaly detected by SFAM. In each of these plots, the difference between the controller output and the steering command corresponding to minimum dissimilarity score is plotted as a function of time. By thresholding on the magnitude of this difference, instantaneous anomalies are detected (red dots) and an overall anomaly detection trigger signal is generated.}
  \label{fig:anomaly_det_plot}
  \vspace*{-0.1in}
\end{figure}

\begin{figure*}[!tpbh]

\centerline{{
\hspace{-0.1cm}
\includegraphics[width=0.115\textwidth, height = 0.12\textwidth]{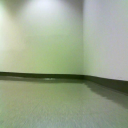}
\includegraphics[width=0.115\textwidth, height = 0.12\textwidth]{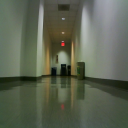}
\includegraphics[width=0.115\textwidth, height = 0.12\textwidth]{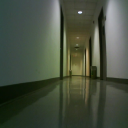}
\includegraphics[width=0.115\textwidth, height = 0.12\textwidth]{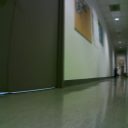}}
{
\includegraphics[width=0.115\textwidth, height = 0.12\textwidth]{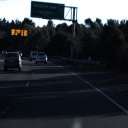}
\includegraphics[width=0.115\textwidth, height = 0.12\textwidth]{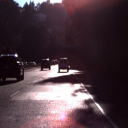}
\includegraphics[width=0.115\textwidth, height = 0.12\textwidth]{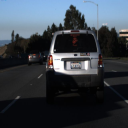}
\includegraphics[width=0.115\textwidth, height = 0.12\textwidth]{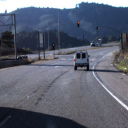}}}

\centerline{\hspace{-0.2cm}{
\includegraphics[width=0.12\textwidth, height = 0.13\textwidth, trim=0 0 1.1cm 0,clip=true]{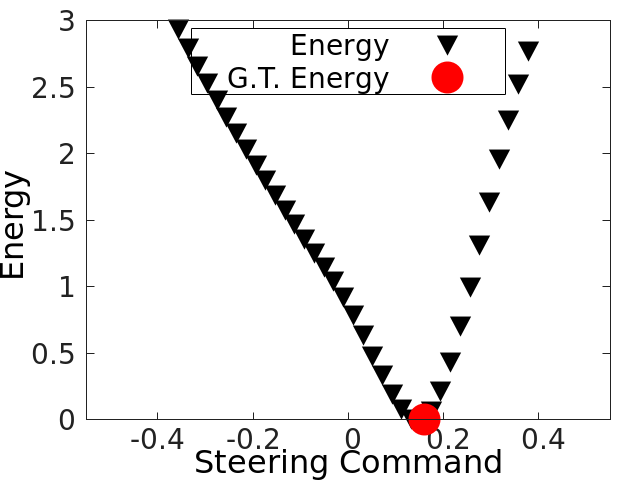}
\includegraphics[width=0.12\textwidth, height = 0.13\textwidth, trim=1.1cm 0 1.1cm 0,clip=true]{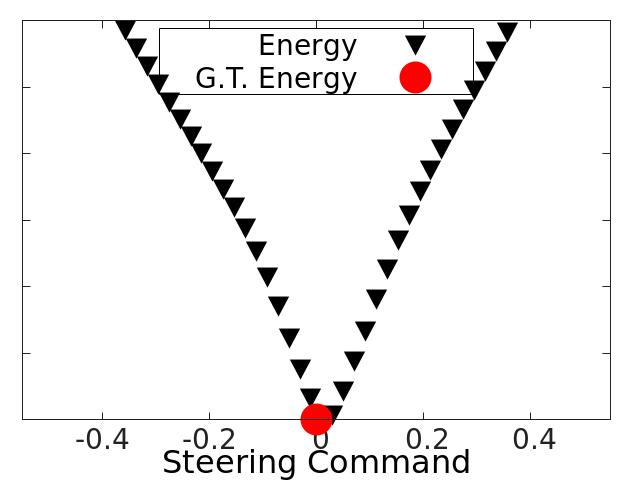}
\includegraphics[width=0.12\textwidth, height = 0.13\textwidth, trim=1.1cm 0 1.1cm 0,clip=true]{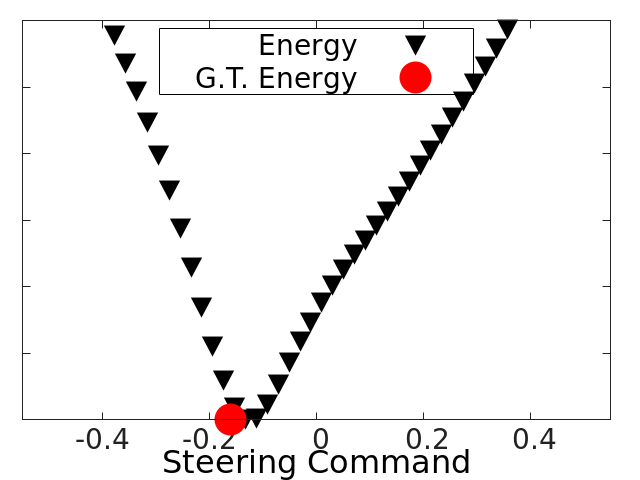}
\includegraphics[width=0.12\textwidth, height = 0.13\textwidth, trim=1.1cm 0 1.1cm 0,clip=true]{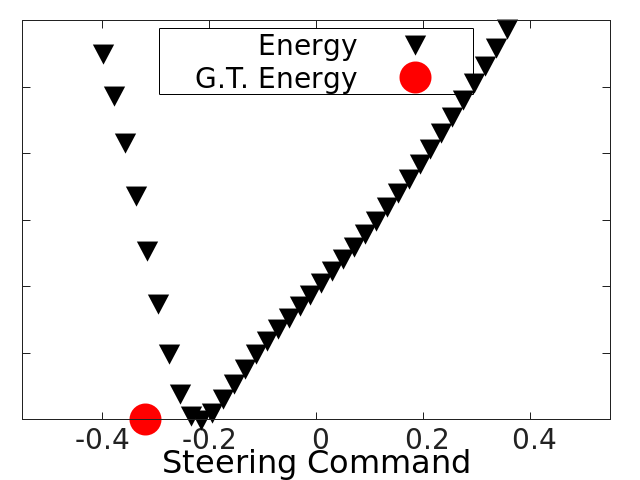}
}
\hspace{-0.4cm} 
{
\includegraphics[width=0.12\textwidth, height = 0.13\textwidth, trim=0 0 1.1cm 0,clip=true]{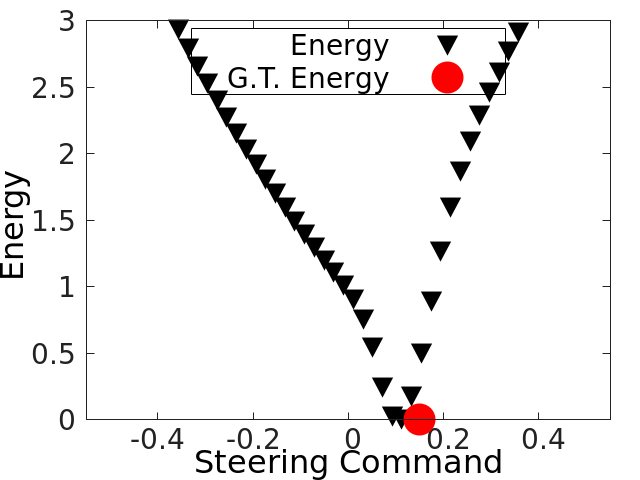}
\includegraphics[width=0.12\textwidth, height = 0.13\textwidth, trim=1.1cm 0 1.1cm 0,clip=true]{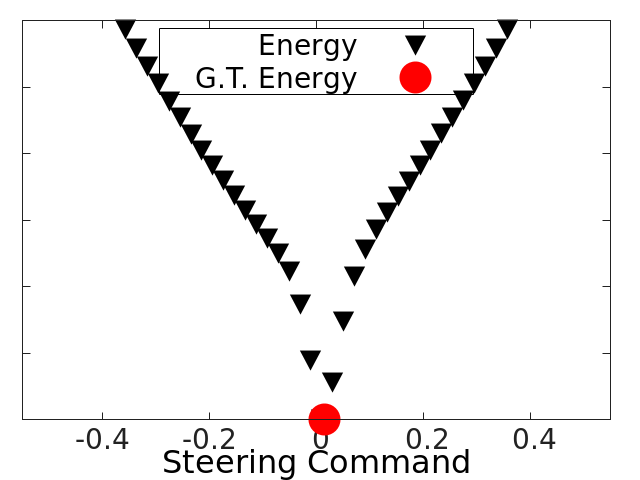}
\includegraphics[width=0.12\textwidth, height = 0.13\textwidth, trim=1.1cm 0 1.1cm 0,clip=true]{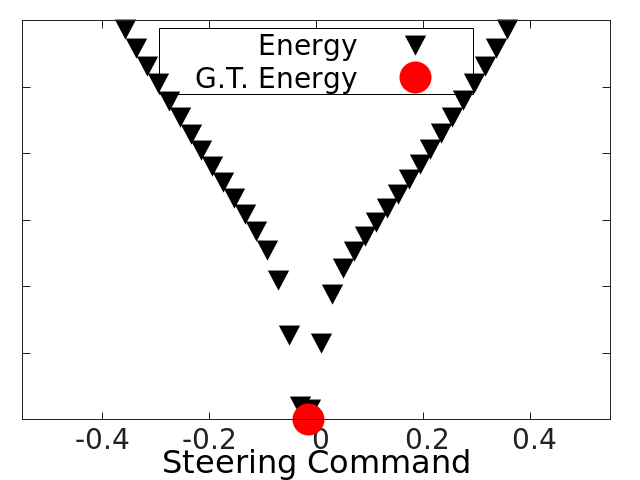}
\includegraphics[width=0.12\textwidth, height = 0.13\textwidth, trim=1.1cm 0 1.1cm 0,clip=true]{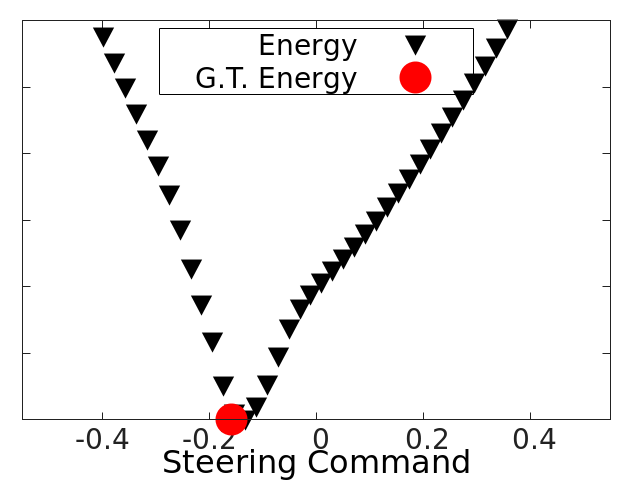}}}

\caption{Examples demonstrating the effectiveness of CFAM for non-anomalous cases in indoor and outdoor environments. In top row, the set of images on the left are from an indoor environment dataset (collected in-house) and the set on the right are from an outdoor environment (Udacity dataset). Each plot in the second row shows the discriminator's outputs plotted as a function of steering command input with corresponding image in top row as condition. It can be observed in the plots that the energy curves have their global minimum close to the ground truth steering command (highlighted as red dots) as expected. Note: These images were taken from the test set and are not used for training the respective discriminators. All three cases of turning left, right, and moving straight are presented.}
\label{fig:CFAM_normal_cases}
\end{figure*}

\begin{figure*}[!tpbh]

\centerline{{
\hspace{0.1cm}
\includegraphics[width=0.115\textwidth, height = 0.12\textwidth]{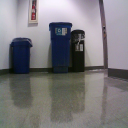}
\includegraphics[width=0.115\textwidth, height = 0.12\textwidth]{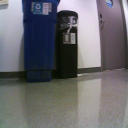}
\includegraphics[width=0.115\textwidth, height = 0.12\textwidth]{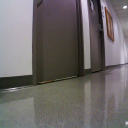}
\includegraphics[width=0.115\textwidth, height = 0.12\textwidth]{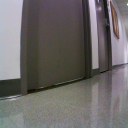}}
{
\includegraphics[width=0.115\textwidth, height = 0.12\textwidth]{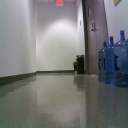}
\includegraphics[width=0.115\textwidth, height = 0.12\textwidth]{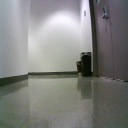}
\includegraphics[width=0.115\textwidth, height = 0.12\textwidth]{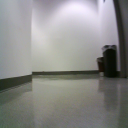}
\includegraphics[width=0.115\textwidth, height = 0.12\textwidth]{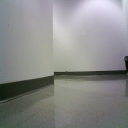}}}

\centerline{{
\includegraphics[width=0.12\textwidth, height = 0.13\textwidth, trim=0 0 1.1cm 0,clip=true]{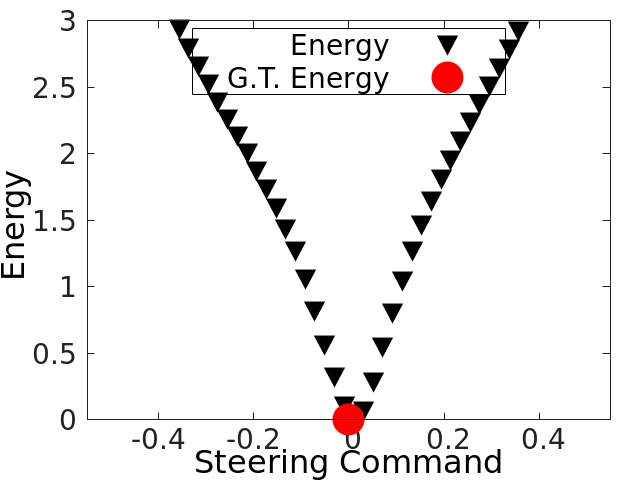}
\includegraphics[width=0.12\textwidth,height = 0.13\textwidth,trim=1.1cm 0 1.1cm 0,clip=true]{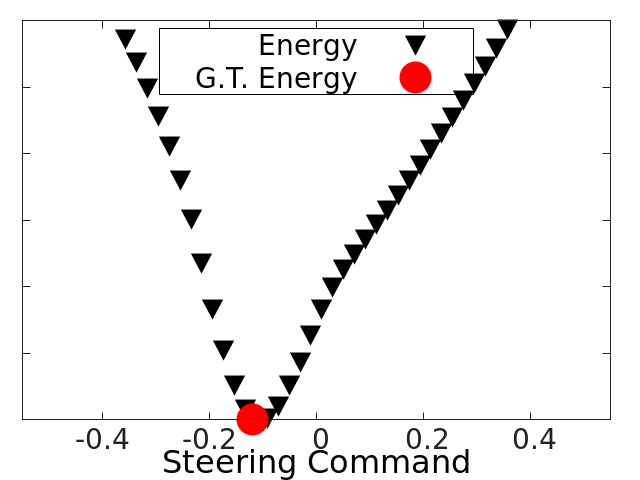}
\includegraphics[width=0.12\textwidth, height = 0.13\textwidth,trim=1.1cm 0 1.1cm 0,clip=true]{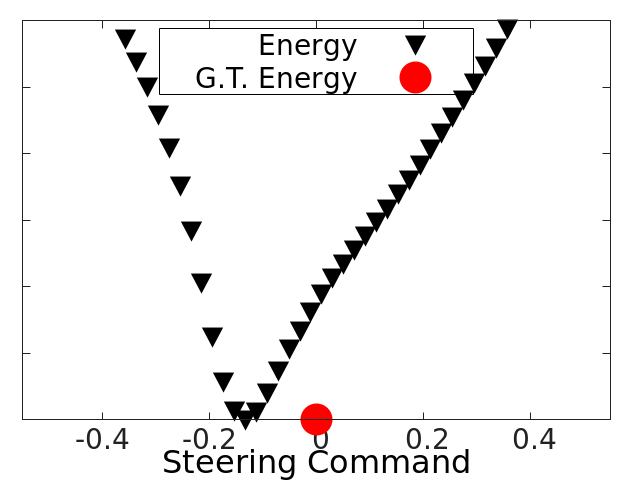}
\includegraphics[width=0.12\textwidth,height = 0.13\textwidth ,trim=1.1cm 0 1.1cm 0,clip=true]{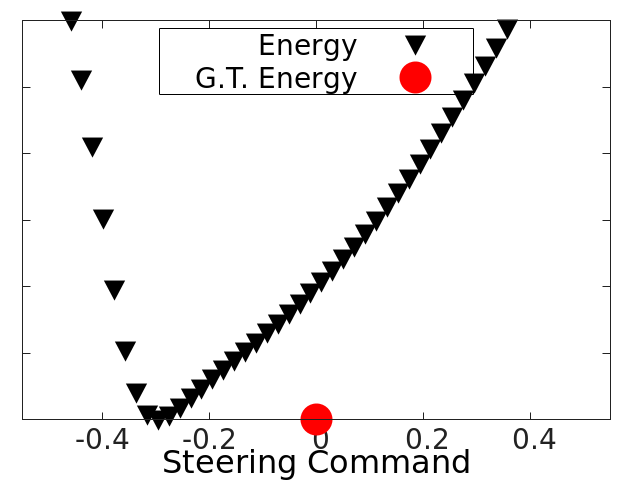}
}
\hspace{-0.4cm}
{
\includegraphics[width=0.12\textwidth, height = 0.13\textwidth, trim=0 0 1.1cm 0,clip=true]{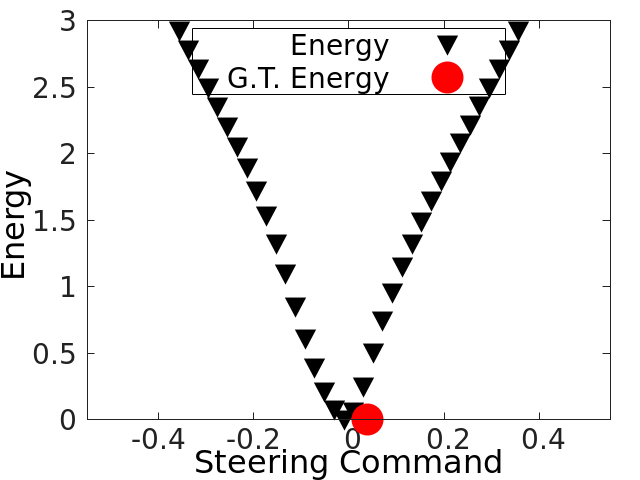}
\includegraphics[width=0.12\textwidth,height = 0.13\textwidth,trim=1.1cm 0 1.1cm 0,clip=true]{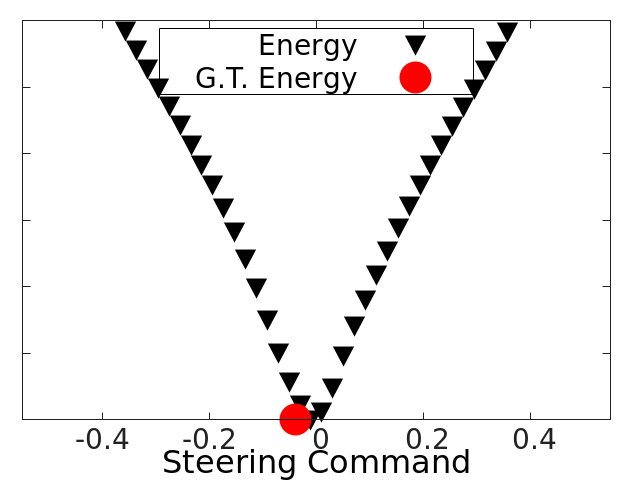}
\includegraphics[width=0.12\textwidth,height = 0.13\textwidth,trim=1.1cm 0 1.1cm 0,clip=true]{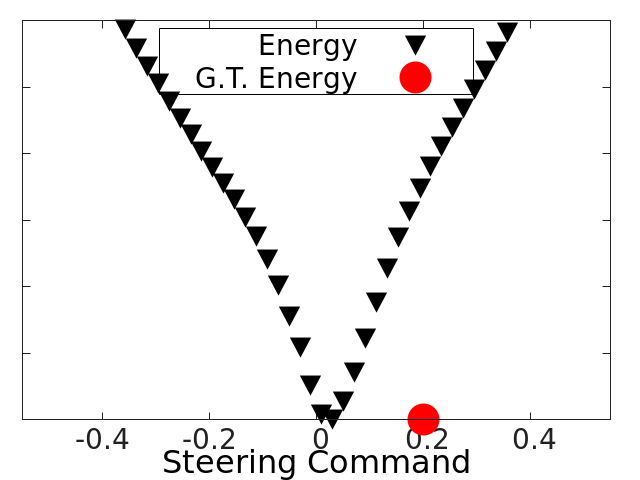}
\includegraphics[width=0.11\textwidth,height = 0.13\textwidth,trim=1.1cm 0 1.1cm 0,clip=true]{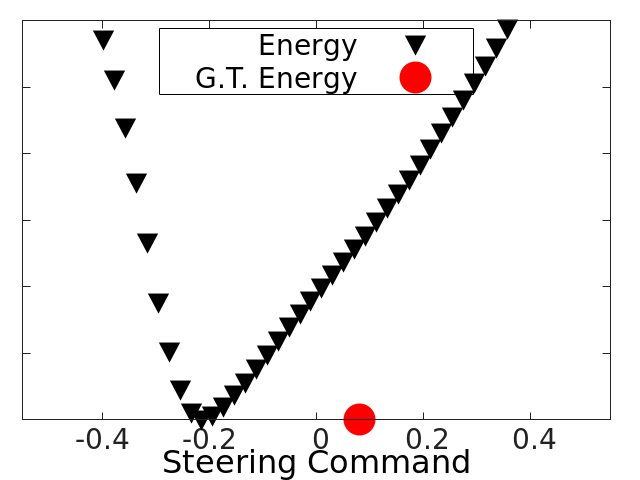}}}

\caption{ Examples demonstrating the effectiveness of CFAM for anomalous cases in indoor environment. The set of images on the left are from an instance when the unmanned ground vehicle (UGV) makes a late right turn and on the right are from an instance when the UGV makes an early left turn. Each plot in second row shows the discriminator's outputs plotted as function of steering command input with corresponding image in top row as condition. It can be observed in the plots that third and fourth images in both sets are anomalous cases and the steering command deviations (distance between red dot and the energy curve global minimum) are high, thus signaling anomalies.}
\label{fig:CFAM_anomalies}
\end{figure*}


\begin{figure*}[!tbph]
\centerline{\includegraphics[width=0.48\textwidth]{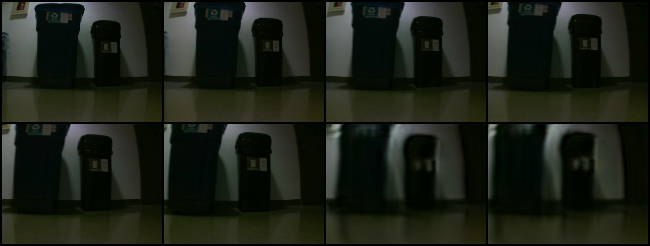}
\includegraphics[width=0.48\textwidth]{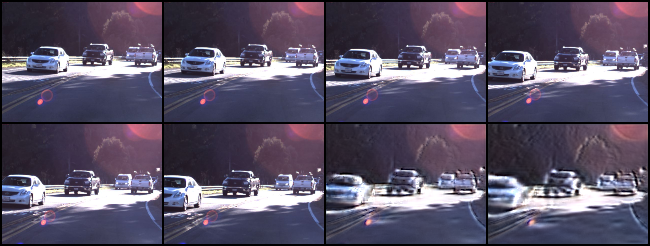}}
    \caption{Temporal prediction of camera images in varying lighting for indoor and outdoor environments. The set of images in the left are from an indoor environment dataset (collected in-house) and the set on the right are from an outdoor environment (Udacity dataset). In each set, the four frames in the top row are the input image sequence given to the temporal prediction network. The first two frames in the bottom row of each set are the two ground-truth future frames and the next two frames are the predicted future frames.}
    \label{fig:videopredindoor}
    \vspace*{-0.2in}
\end{figure*}

\begin{figure*}[!tbph]
\centerline{{
\includegraphics[width=0.118\textwidth]{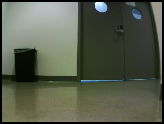}
\includegraphics[width=0.118\textwidth]{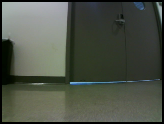}
\includegraphics[width=0.118\textwidth]{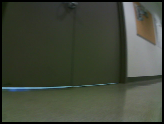}
\includegraphics[width=0.118\textwidth]{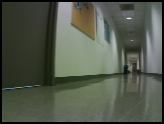}}
{
\includegraphics[width=0.118\textwidth]{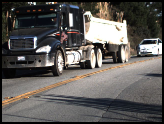}
\includegraphics[width=0.118\textwidth]{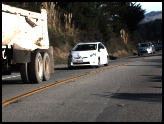}
\includegraphics[width=0.118\textwidth]{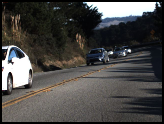}
\includegraphics[width=0.118\textwidth]{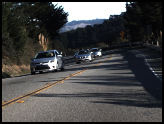}}}

\centerline{{
\includegraphics[width=0.118\textwidth]{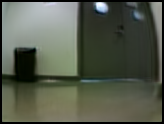}
\includegraphics[width=0.118\textwidth]{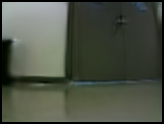}
\includegraphics[width=0.118\textwidth]{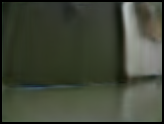}
\includegraphics[width=0.118\textwidth]{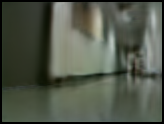}}
{
\includegraphics[width=0.118\textwidth]{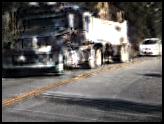}
\includegraphics[width=0.118\textwidth]{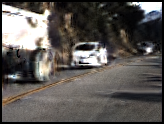}
\includegraphics[width=0.118\textwidth]{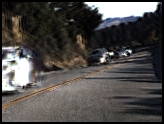}
\includegraphics[width=0.118\textwidth]{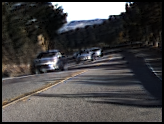}}}

\centerline{{
\includegraphics[width=0.12\textwidth, height=1.83cm]{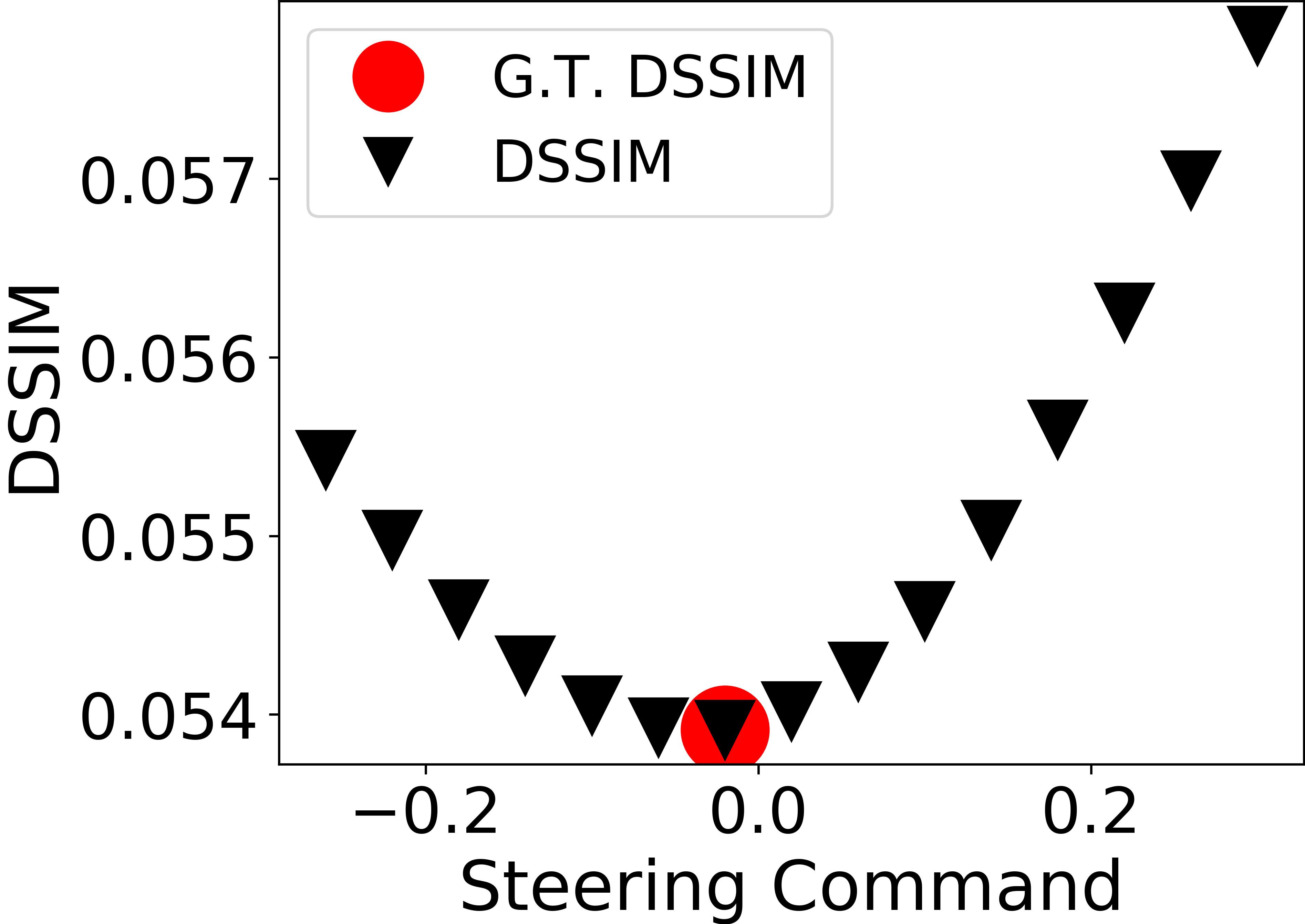}
\includegraphics[width=0.12\textwidth,trim=3.3cm 0 0 0,clip=true]{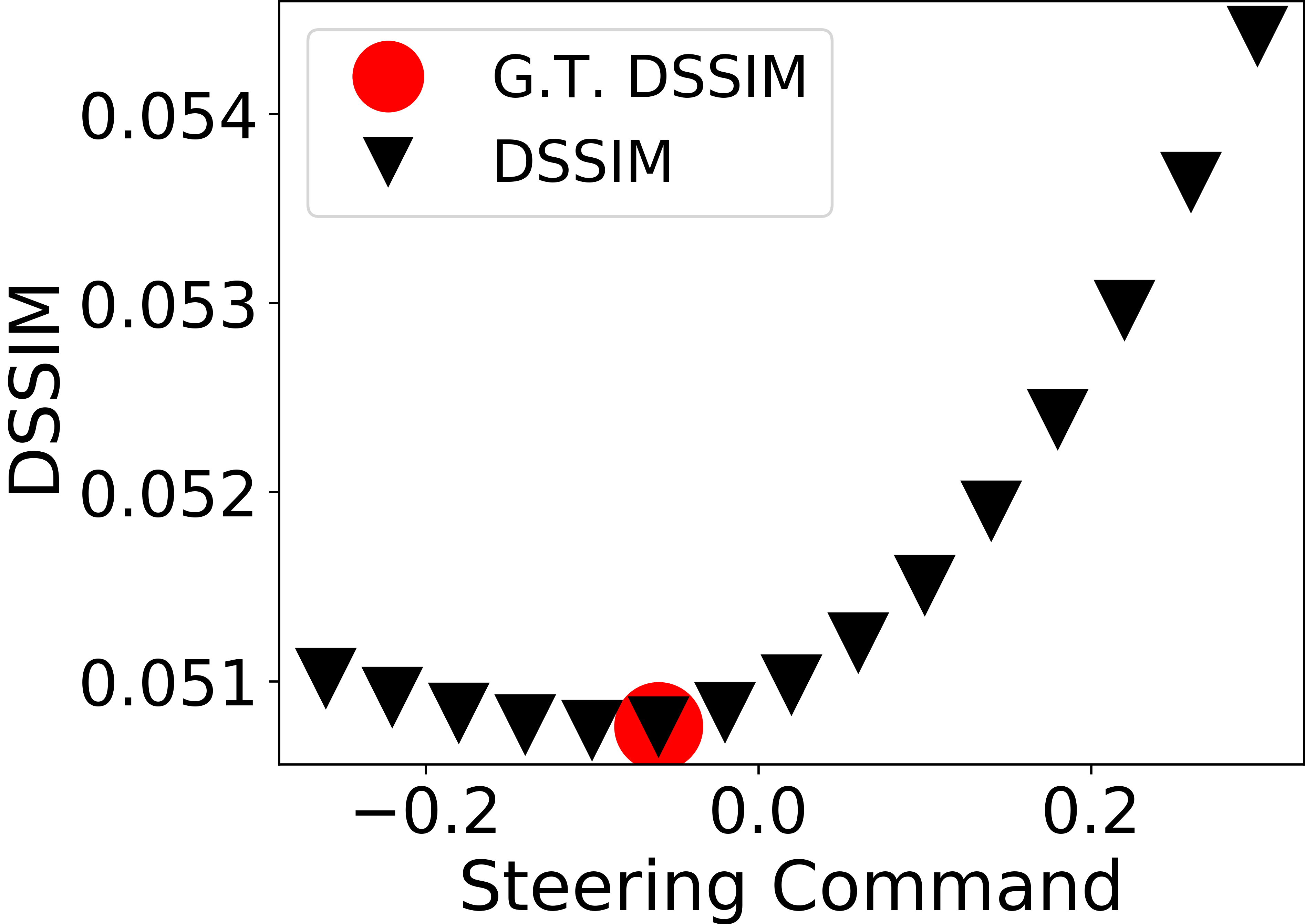}
\includegraphics[width=0.12\textwidth,trim=3.7cm 0 0 0,clip=true]{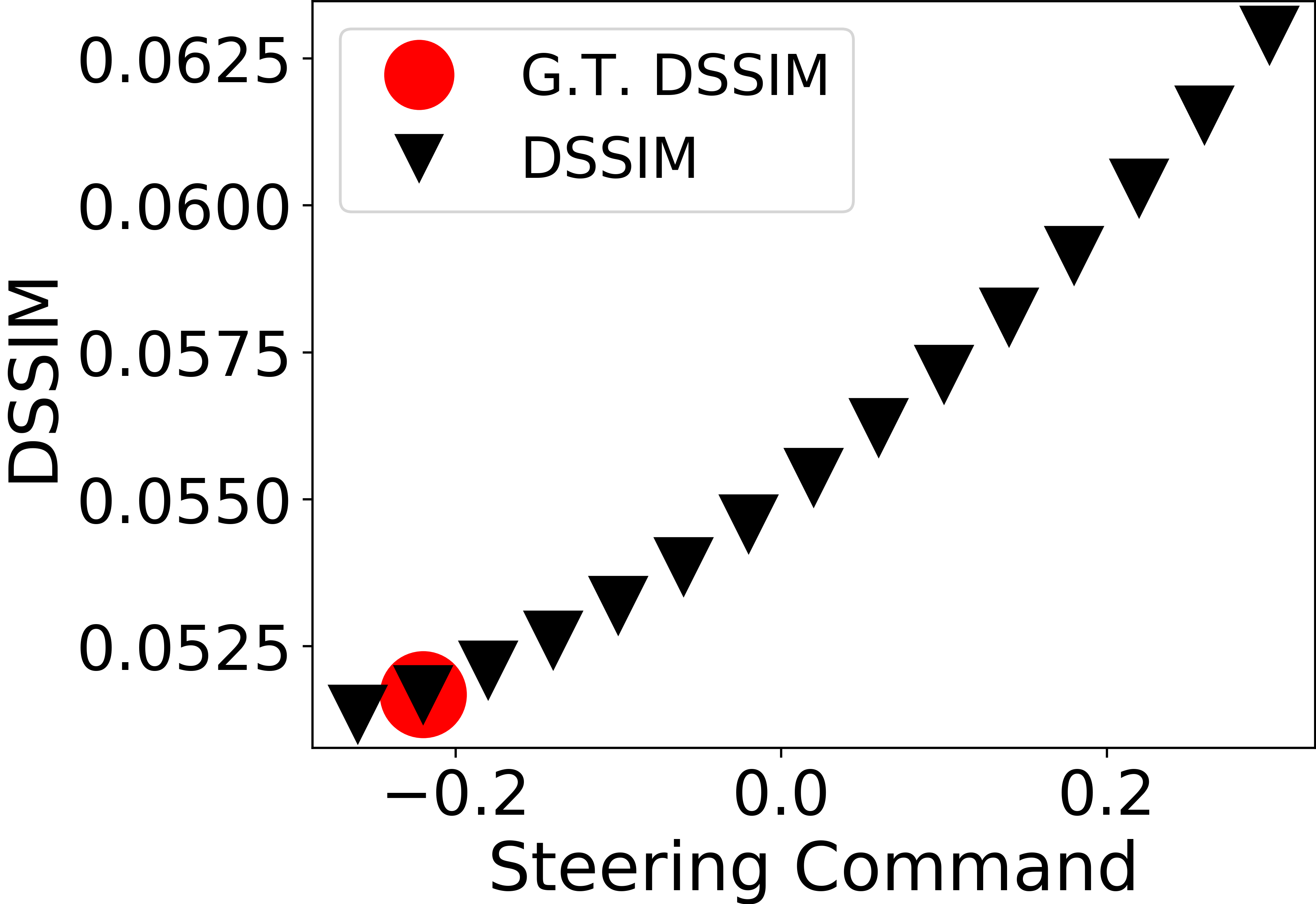}
\includegraphics[width=0.12\textwidth,trim=3.7cm 0 0 0,clip=true]{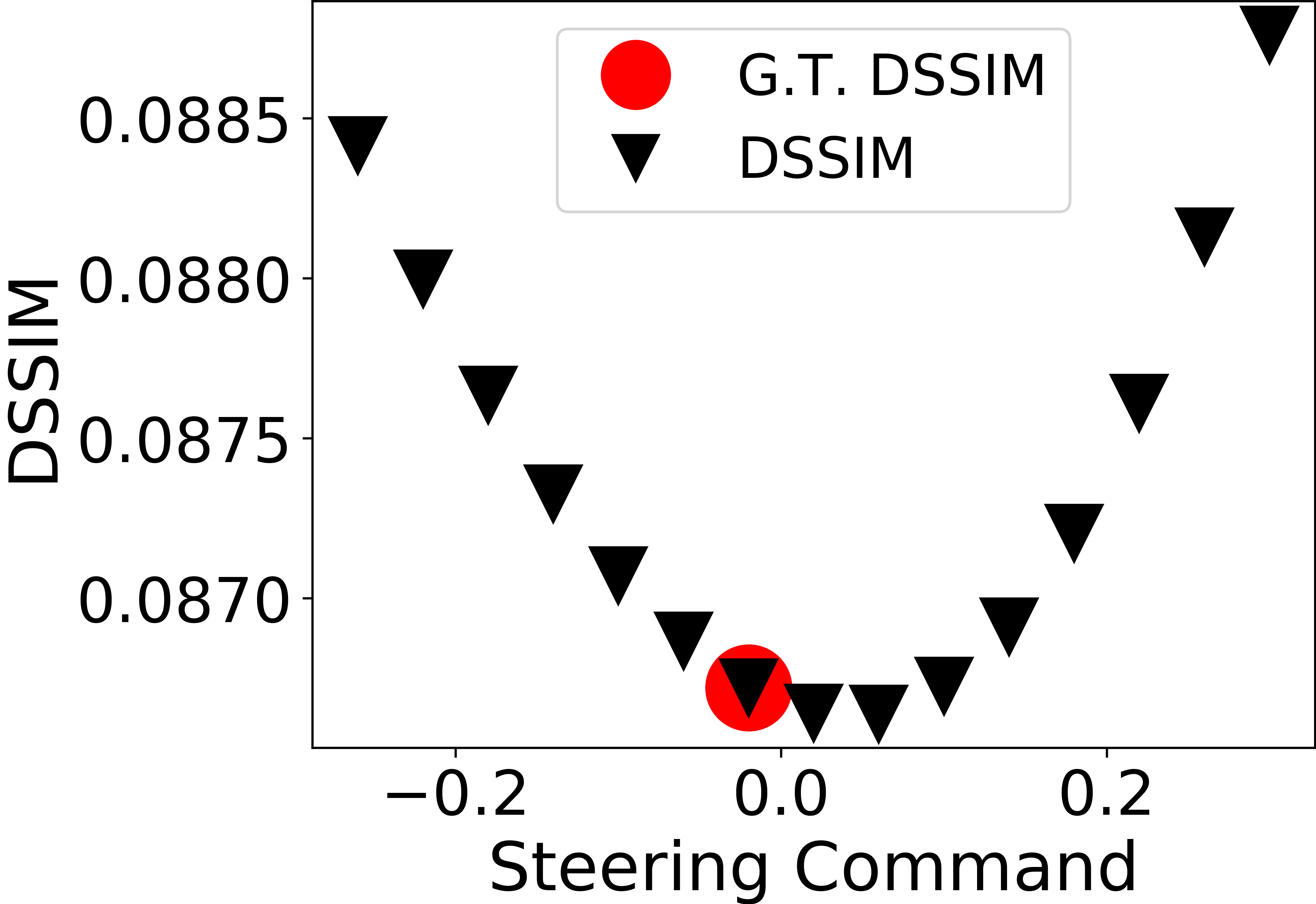}
}
\hspace{-0.35cm}
{
\includegraphics[width=0.12\textwidth,height=1.83cm,trim=1cm 0 0 0,clip=true]{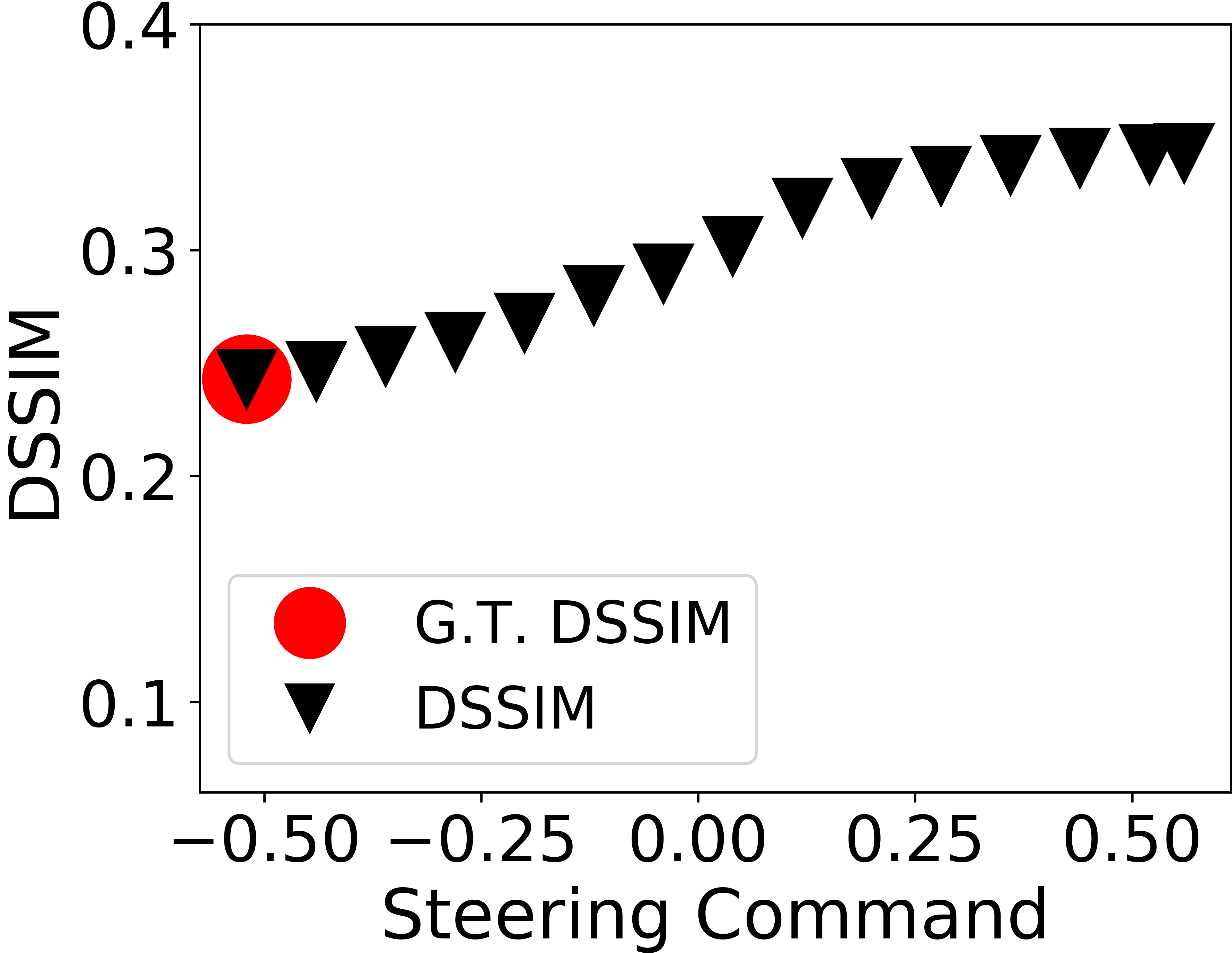}
\includegraphics[width=0.12\textwidth,trim=2.3cm 0 0 0,clip=true]{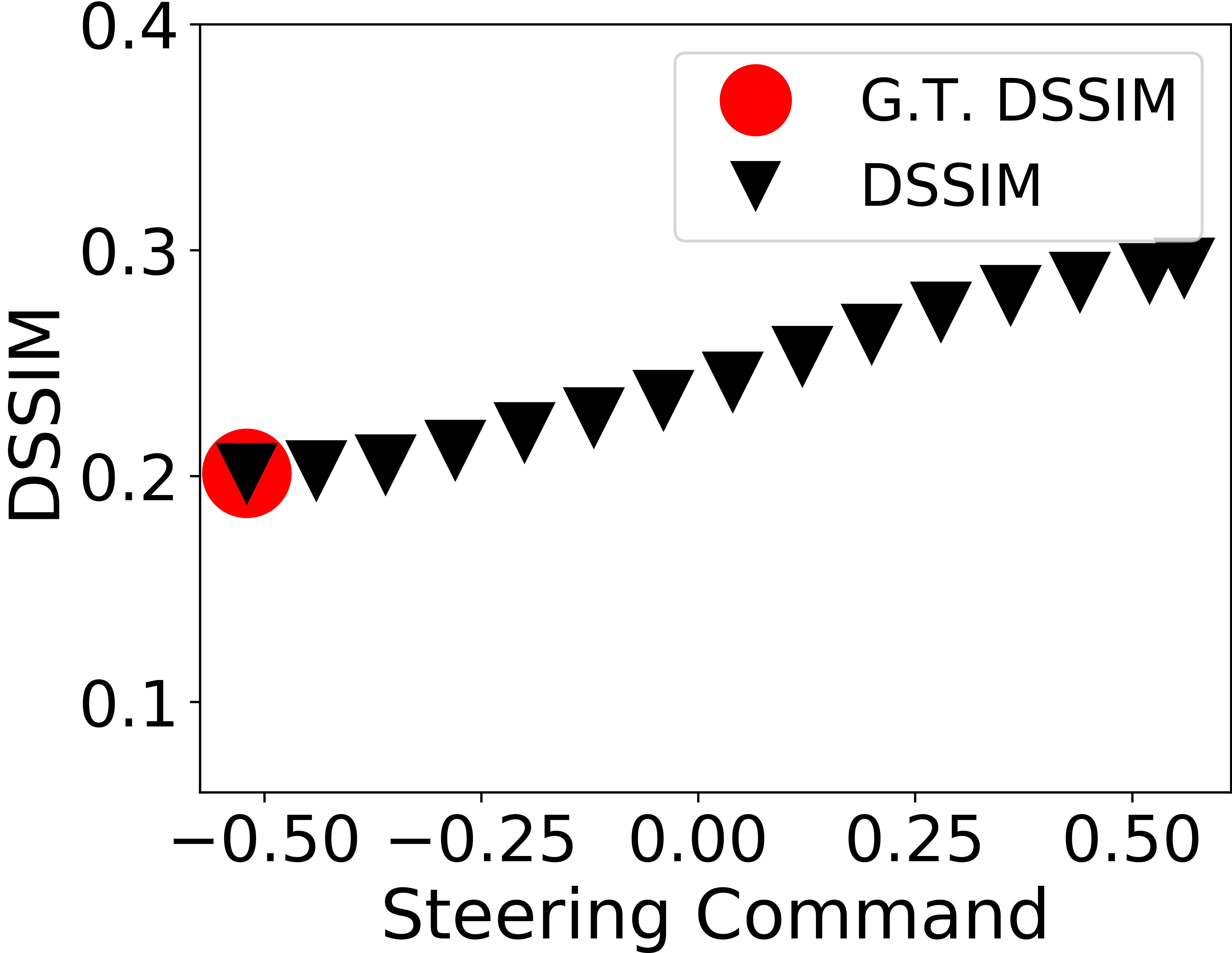}
\includegraphics[width=0.12\textwidth,trim=2.3cm 0 0 0,clip=true]{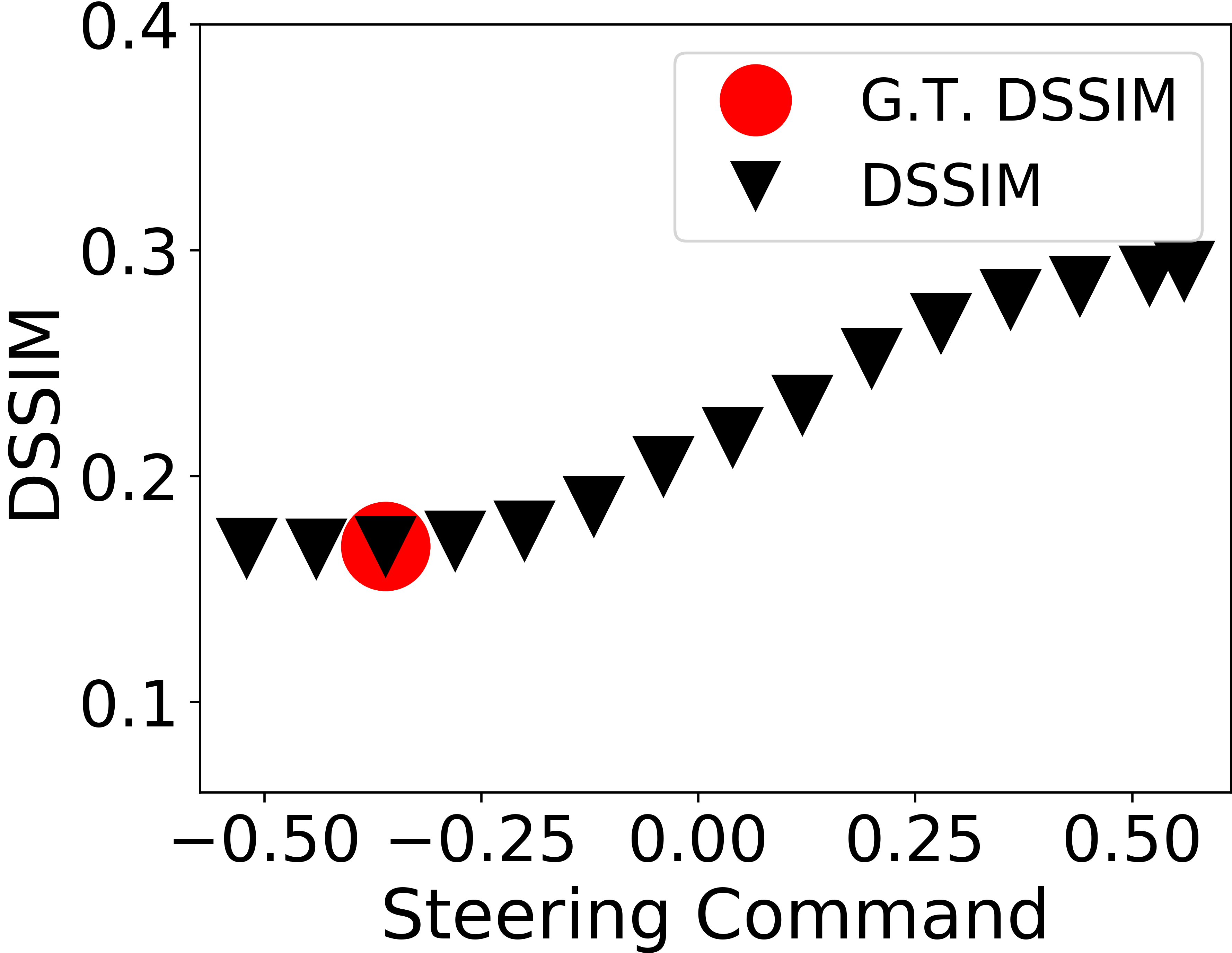}
\includegraphics[width=0.12\textwidth,trim=2.3cm 0 0 0,clip=true]{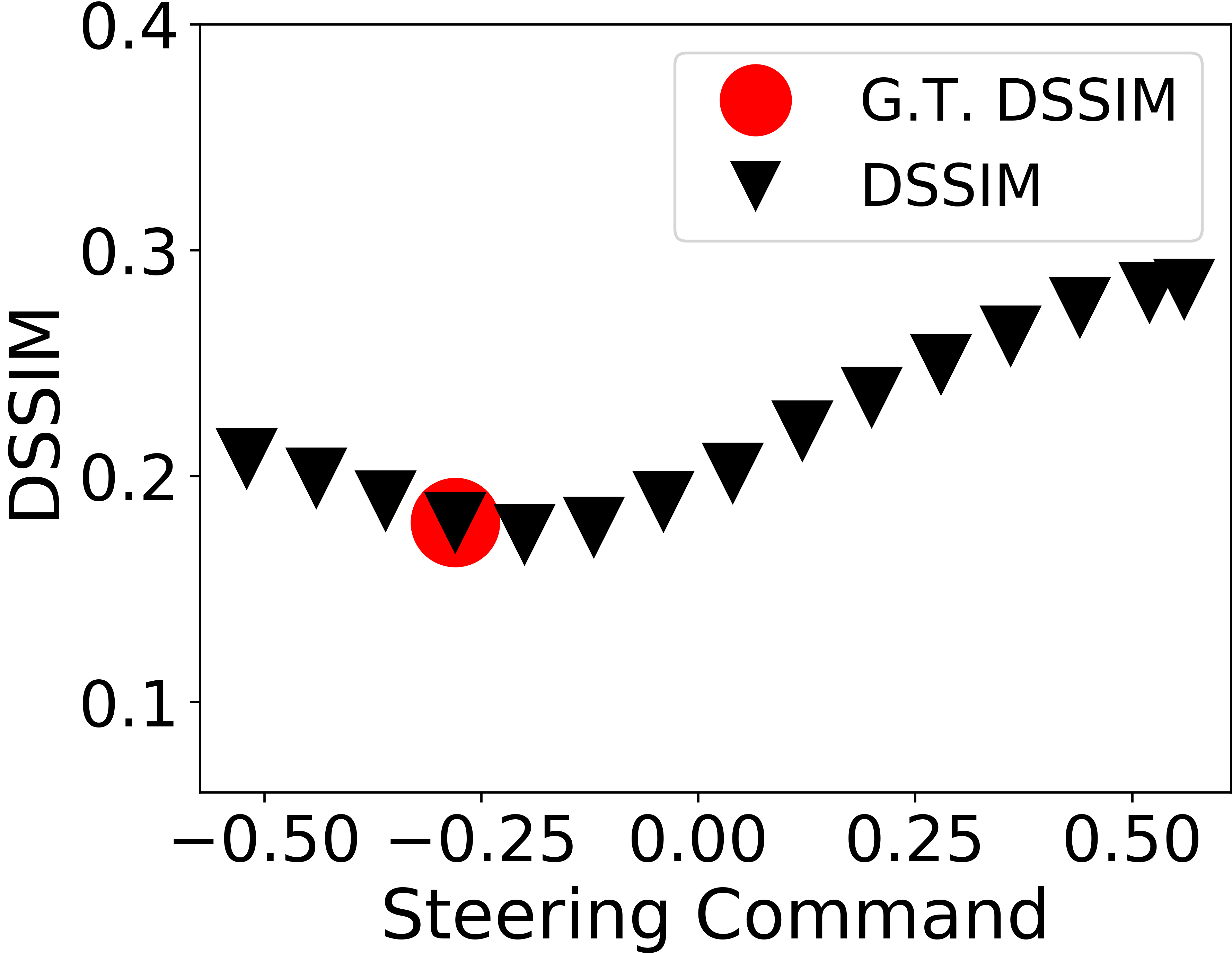}}}

    \caption{Examples demonstrating the effectiveness of the SFAM for non-anomalous cases in indoor and outdoor environments. The set of images in the left are from an indoor environment dataset (collected in-house) and the set on the right are from an outdoor environment (Udacity dataset). In each set, the images on the top are the ground truth of the frame predictions shown in the middle row of each set which are generated conditioned on the actual future action. The last row shows the plot of dissimilarity (DSSIM) ((1-SSIM)/2) between the ground truth future frame and frame predictions conditioned on 15 different actions (same number of actions, as used in the discriminator). The steering command corresponding to action conditioned predicted frame with the lowest dissimilarity is selected and compared with the actual steering command to detect an anomaly. It can be inferred from the plots that the given cases are non-anomalous.}
    \label{fig:videoprednorm_indoor}
\end{figure*}

\begin{figure*}[!tbph]

\centerline{{
\includegraphics[width=0.118\textwidth]{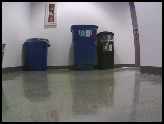}
\includegraphics[width=0.118\textwidth]{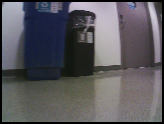}
\includegraphics[width=0.118\textwidth]{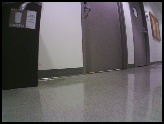}
\includegraphics[width=0.118\textwidth]{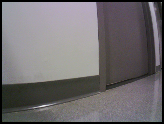}}
{
\includegraphics[width=0.118\textwidth]{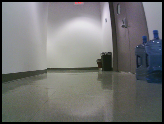}
\includegraphics[width=0.118\textwidth]{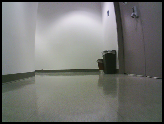}
\includegraphics[width=0.118\textwidth]{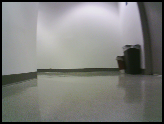}
\includegraphics[width=0.118\textwidth]{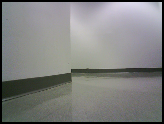}}}

\centerline{{
\includegraphics[width=0.118\textwidth]{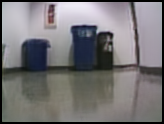}
\includegraphics[width=0.118\textwidth]{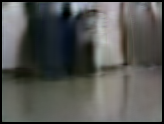}
\includegraphics[width=0.118\textwidth]{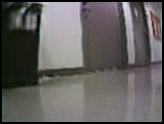}
\includegraphics[width=0.118\textwidth]{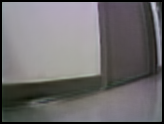}}
{
\includegraphics[width=0.118\textwidth]{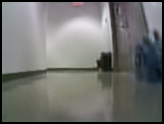}
\includegraphics[width=0.118\textwidth]{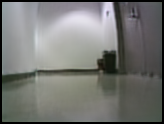}
\includegraphics[width=0.118\textwidth]{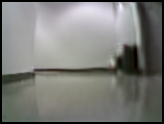}
\includegraphics[width=0.118\textwidth]{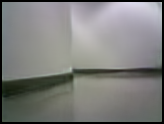}}}

\centerline{{
\includegraphics[width=0.12\textwidth,height=1.83cm]{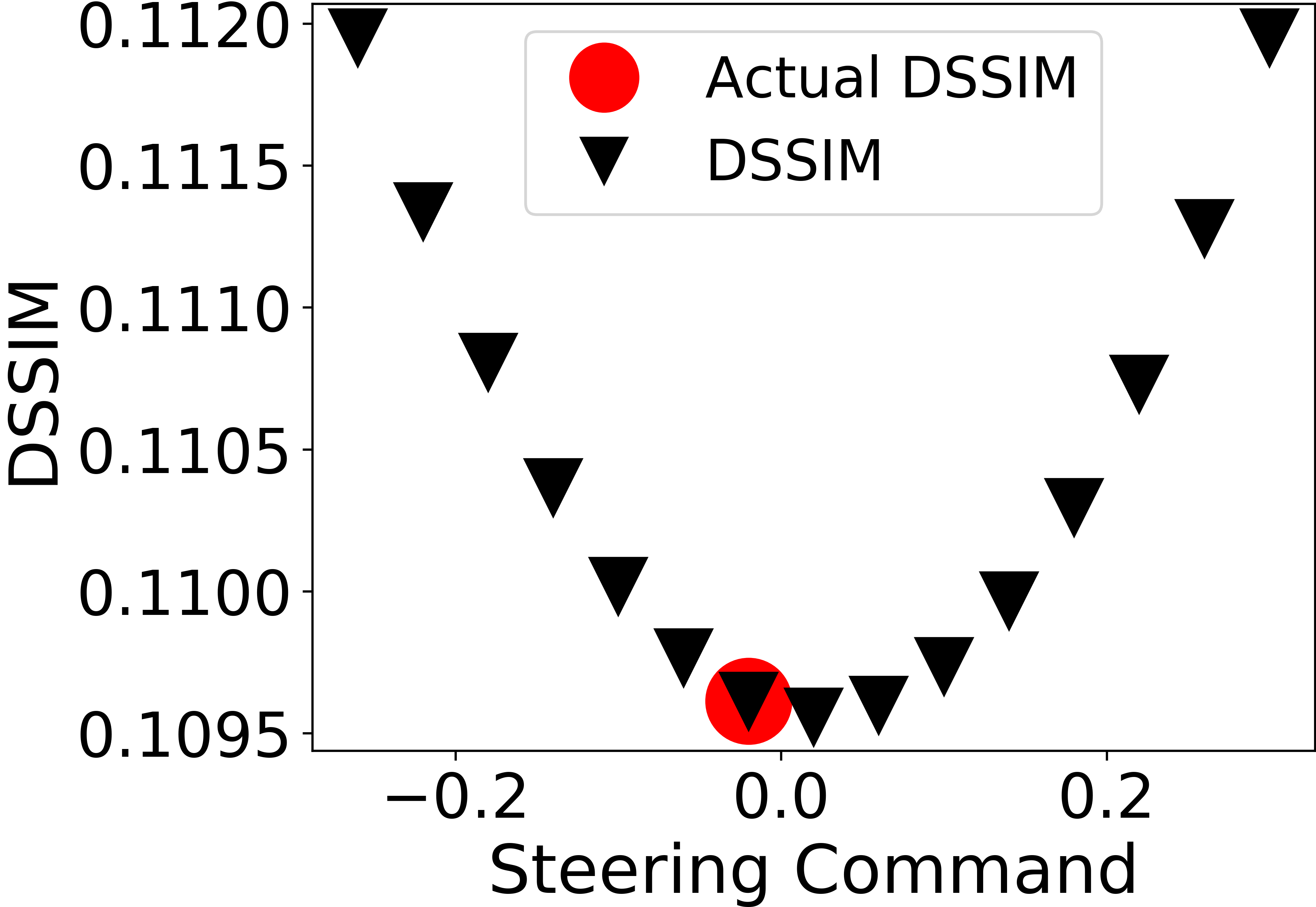}
\includegraphics[width=0.12\textwidth,trim=3.3cm 0 0 0,clip=true]{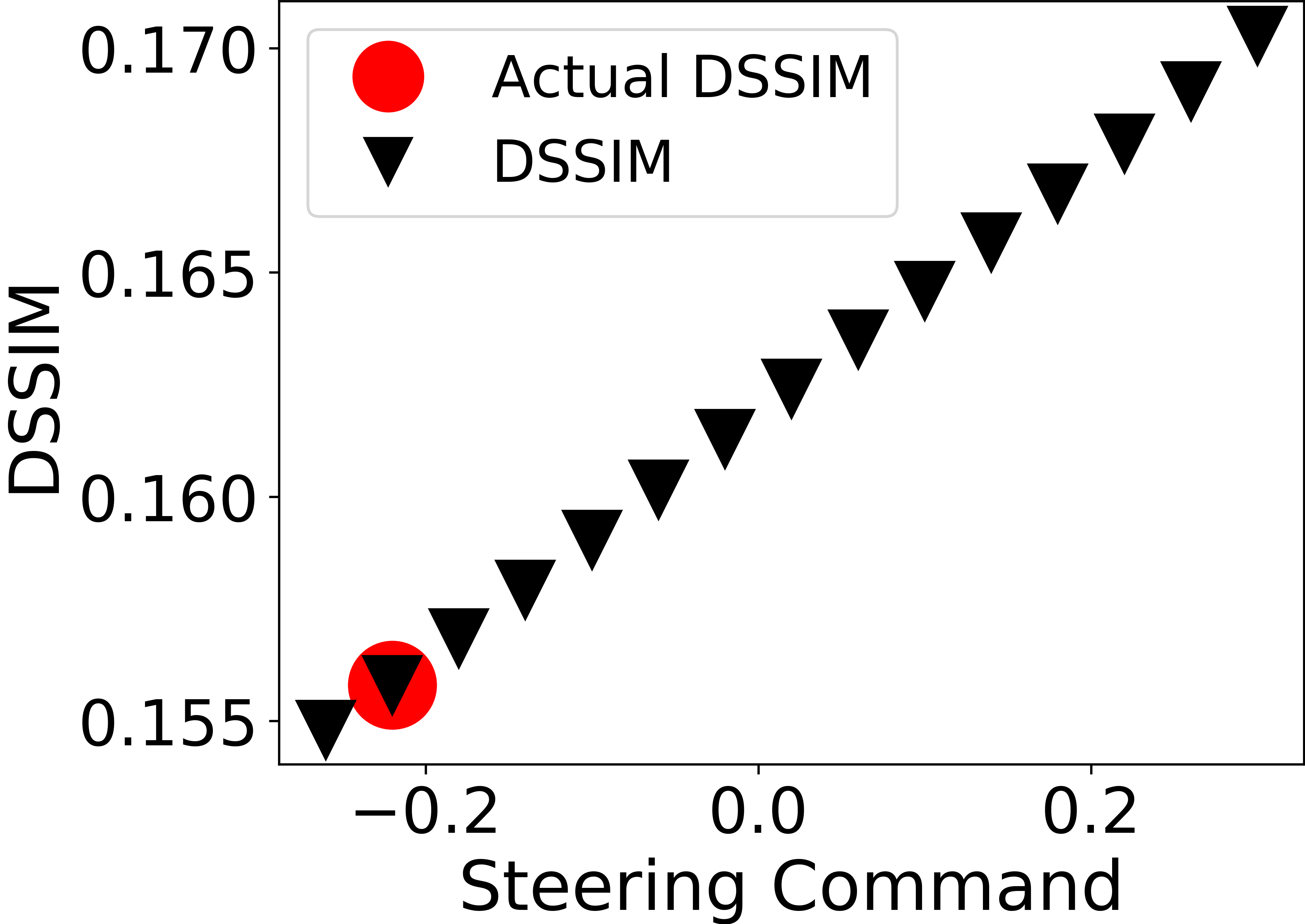}
\includegraphics[width=0.12\textwidth,trim=3.7cm 0 0 0,clip=true]{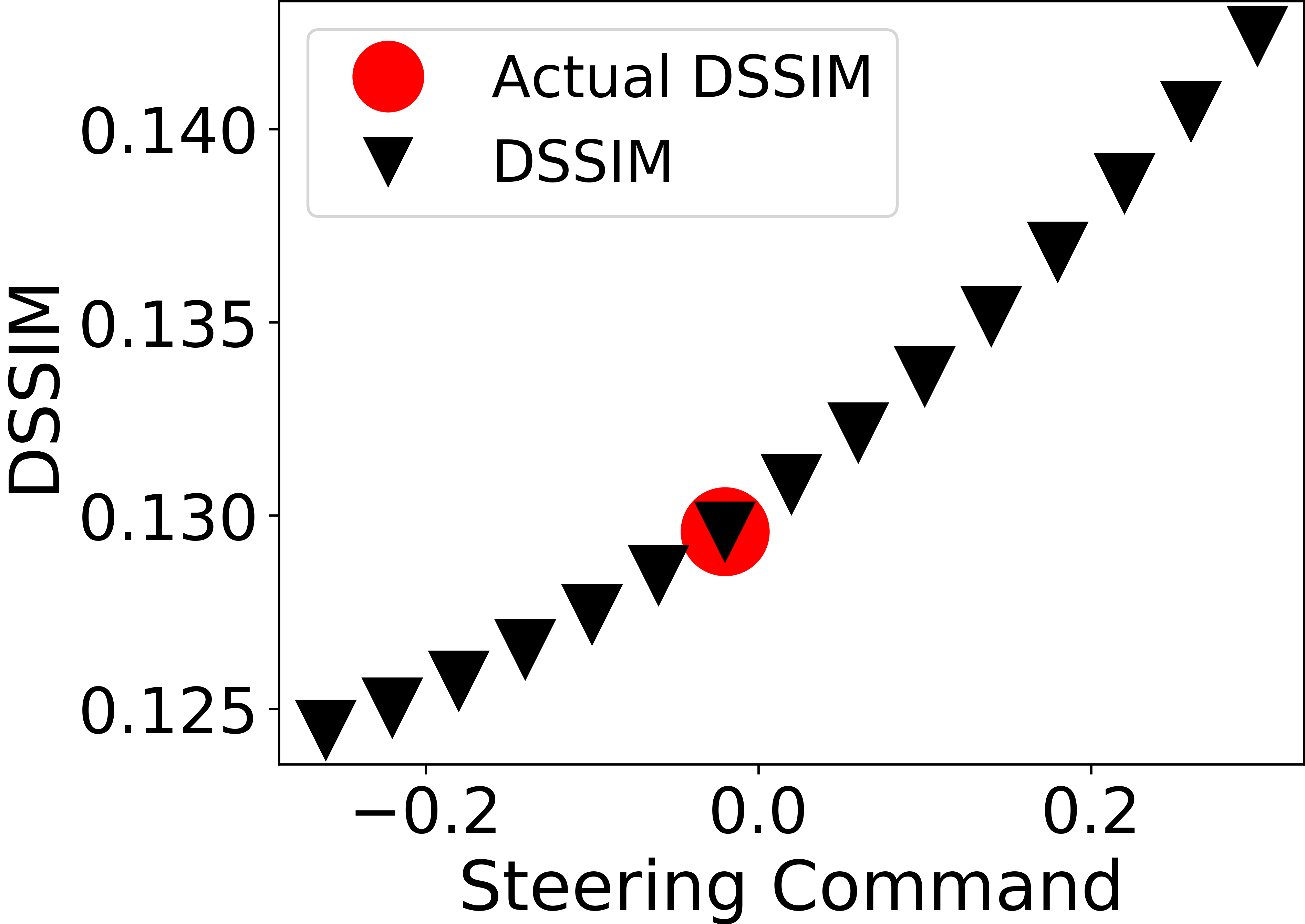}
\includegraphics[width=0.12\textwidth,trim=3.7cm 0 0 0,clip=true]{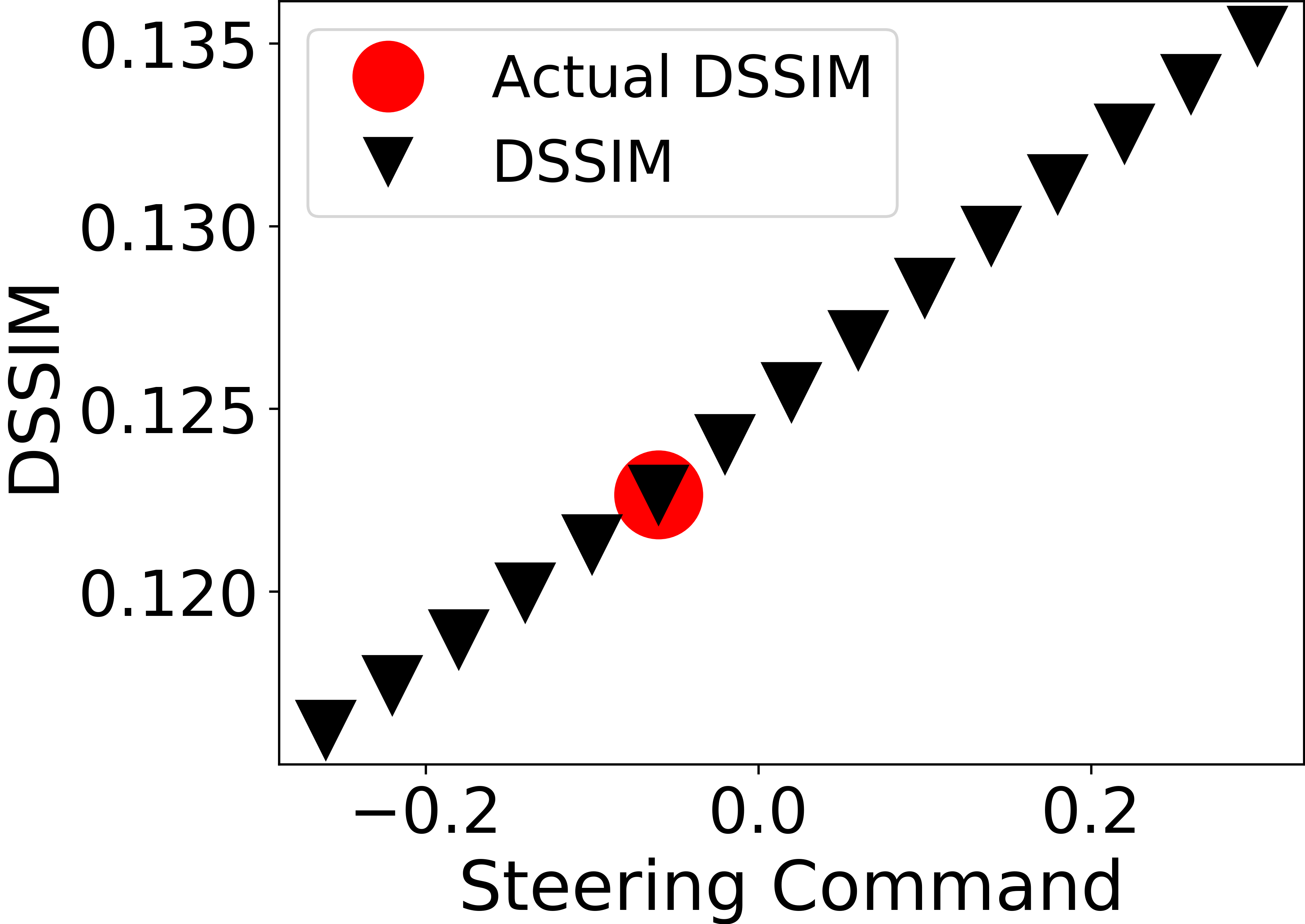}
}
\hspace{-0.35cm}
{
\includegraphics[width=0.12\textwidth,height=1.83cm,trim=1cm 0 0 0,clip=true]{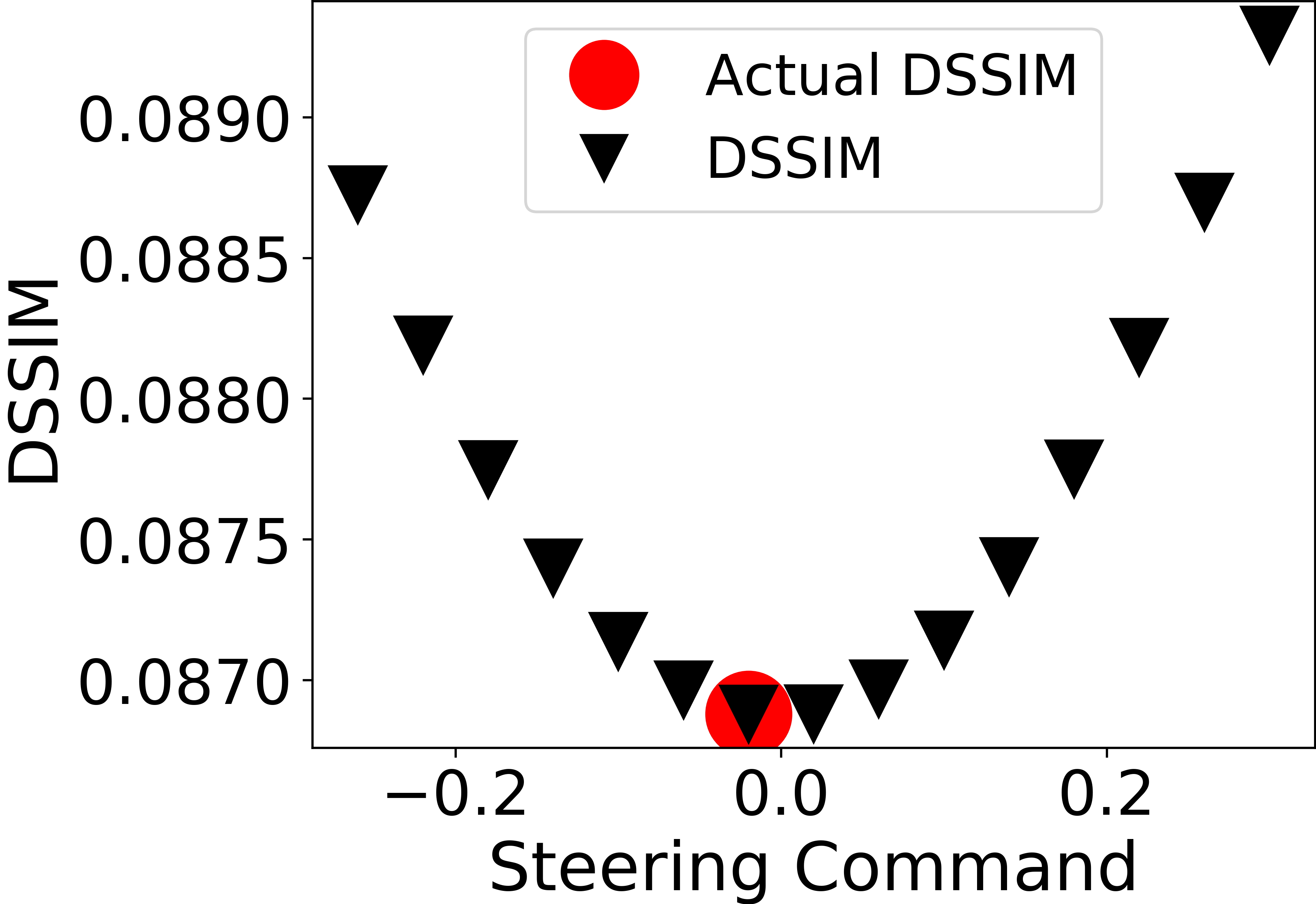}
\includegraphics[width=0.12\textwidth,trim=3.3cm 0 0 0,clip=true]{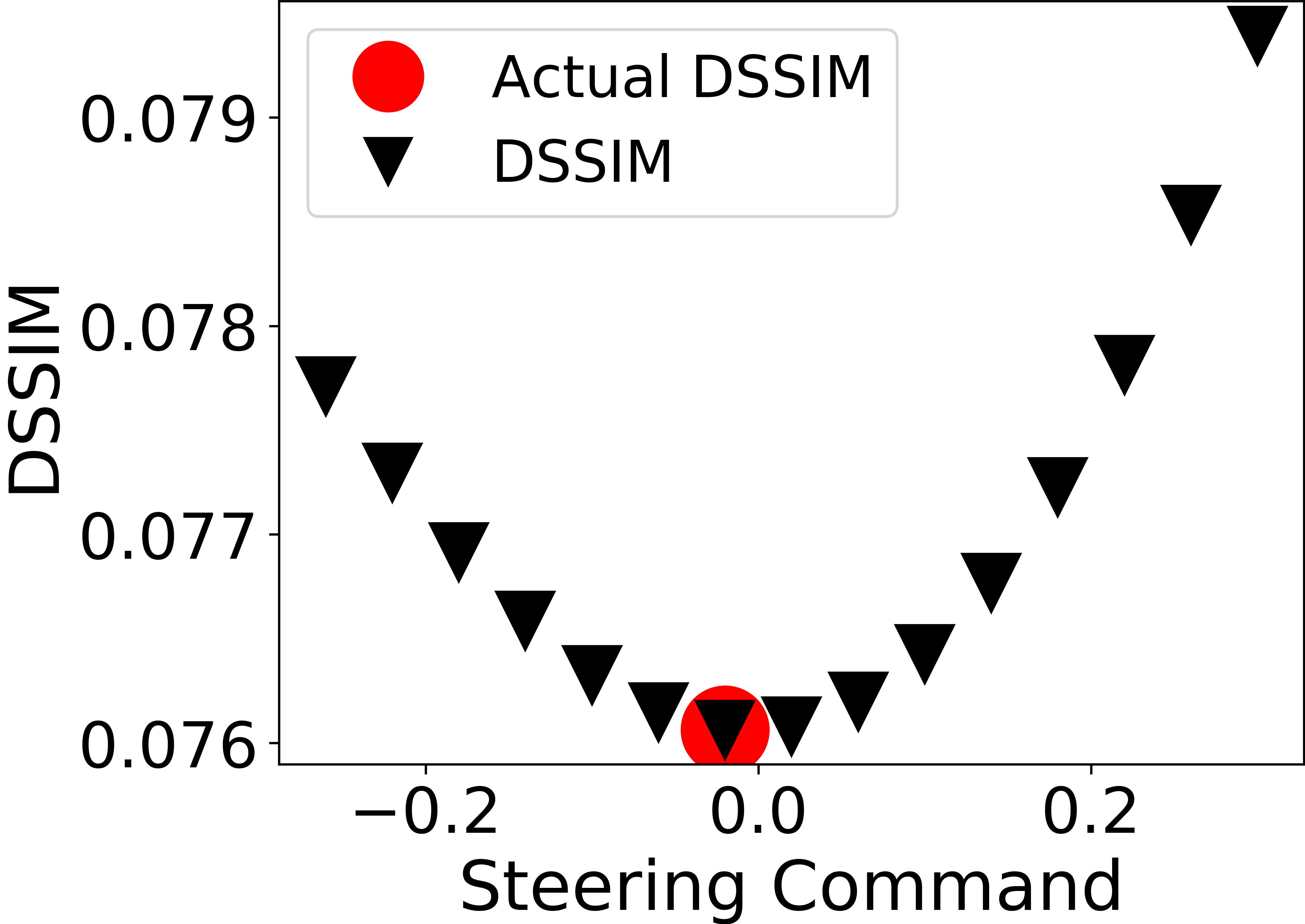}
\includegraphics[width=0.12\textwidth,trim=3.5cm 0 0 0,clip=true]{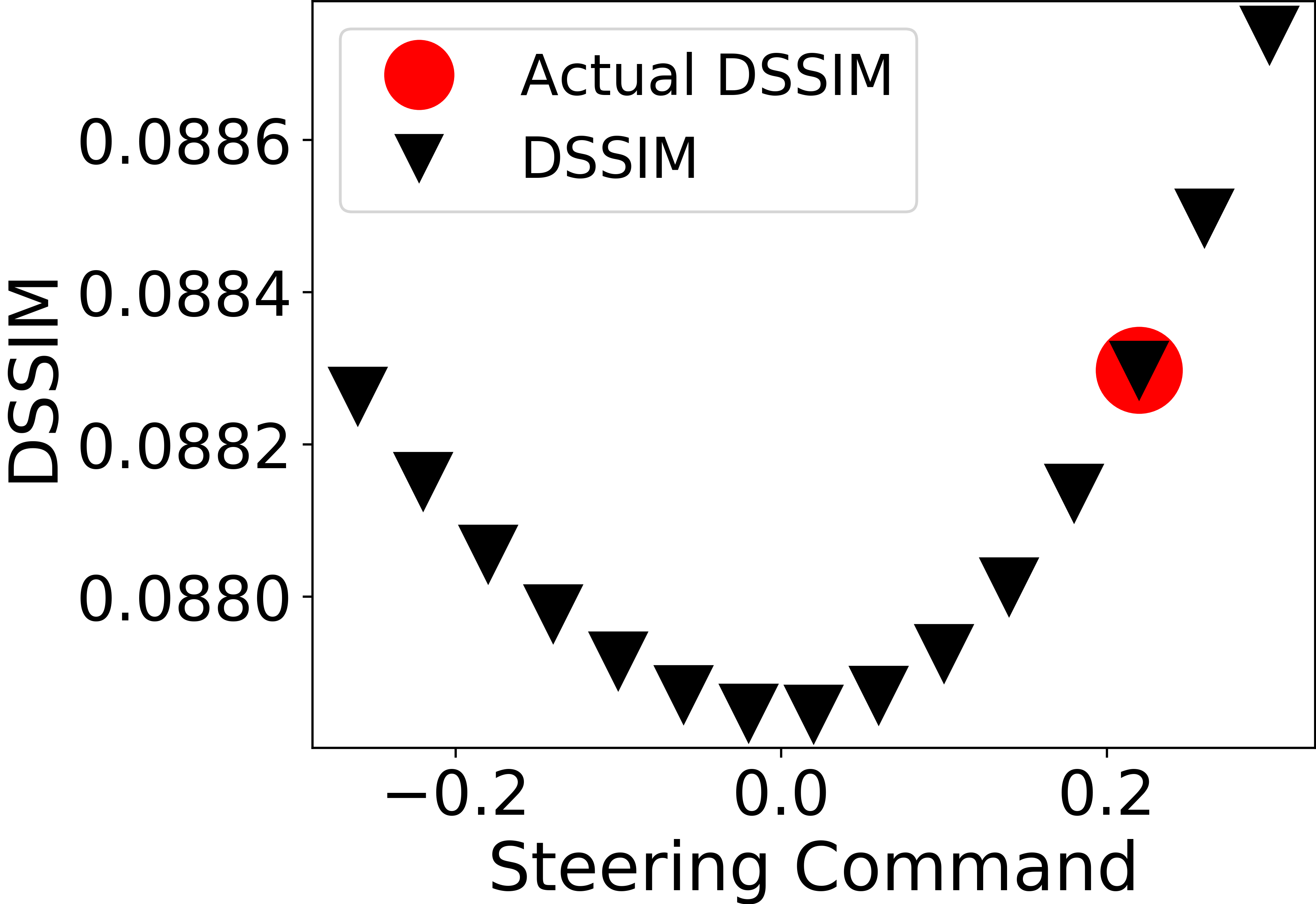}
\includegraphics[width=0.12\textwidth,trim=3.3cm 0 0 0,clip=true]{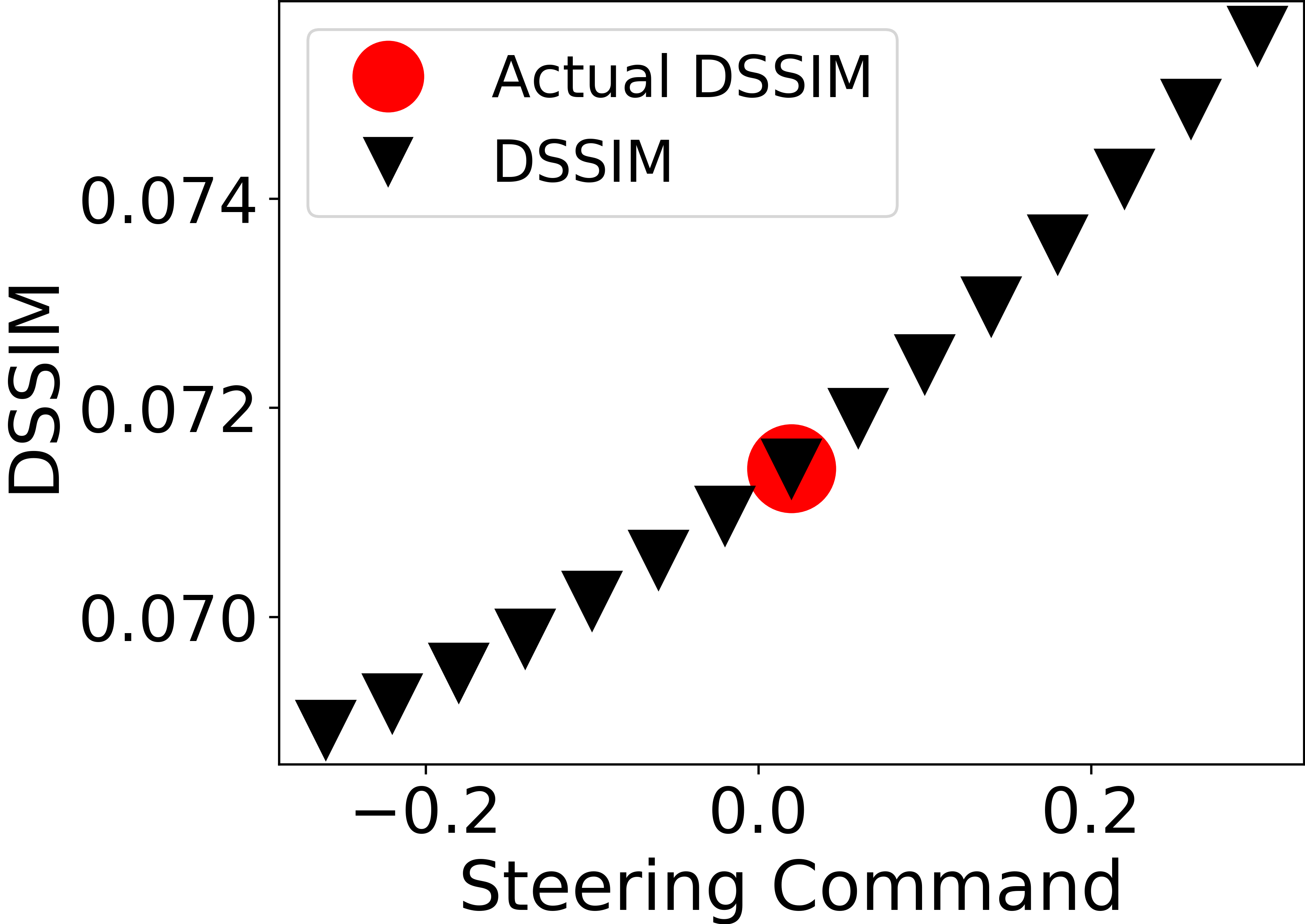}}}
    \caption{Examples demonstrating the effectiveness of SFAM for anomalous cases in indoor environment. The set of images in the left are from an instance when the UGV takes a late right turn and on the right are from an instance when the UGV takes an early left turn. In each set, the images on the top are the ground truth of the frame predictions shown in the middle row of each set which are generated conditioned on the actual future action. The last row shows the plot of dissimilarity (DSSIM) ((1-SSIM)/2) between the ground truth future frame and frame predictions conditioned on 15 different actions (same number of actions, as used in the discriminator). The steering command corresponding to action conditioned predicted frame with the lowest dissimilarity is selected and compared with the actual steering command to detect an anomaly. It is seen that the third and fourth images on both the sets are anomalous since the predicted action conditioned frame does not provide the best match to the actual action conditioned frame.}
    \label{fig:videopredanom}
    \vspace*{-0.22in}
\end{figure*}

CFAM and SFAM performance for non-anomalous scenarios are shown in Figures \ref{fig:CFAM_normal_cases} and \ref{fig:videoprednorm_indoor}, respectively and for anomalous scenarios in Figures \ref{fig:CFAM_anomalies} and \ref{fig:videopredanom}, respectively. The predictions generated by the video prediction architecture in SFAM with varying lighting are shown in Figure \ref{fig:videopredindoor}. By using a threshold on the deviation between the actual steering command and the steering command corresponding to minimum dissimilarity score, a boolean value of anomalous/non-anomalous can be generated as shown in Figure~\ref{fig:anomaly_det_plot}.

\section{CONCLUSIONS}
\label{sec:conclusion}
Our proposed framework is successfully able to continuously validate the mappings from sensor data to actuator command and actuator command to (future) sensor data to detect anomalies/malfunctions in the overall system.


\renewcommand{\baselinestretch}{0.97}
\bibliographystyle{IEEEtran}
\bibliography{refs}

\end{document}